\documentclass[journal]{IEEEtran}
\usepackage{amsmath,amsfonts}
\usepackage{algpseudocode}
\usepackage{algorithm}
\usepackage{array}
\usepackage[caption=false,font=normalsize,labelfont=sf,textfont=sf]{subfig}
\usepackage{textcomp}
\usepackage{stfloats}
\usepackage{url}
\usepackage{verbatim}
\usepackage{graphicx}
\usepackage{cite}
\usepackage{times}
\usepackage{epsfig}
\usepackage{graphicx}
\usepackage{amssymb}
\usepackage{fontawesome}
\usepackage{multirow}
\usepackage{tabularx}
\usepackage{color}
\usepackage{fancyhdr}
\usepackage{mathtools}
\usepackage{amsthm}
\usepackage[table]{xcolor}
\usepackage{pifont}
\usepackage{algpseudocode}

\definecolor{myblue}{RGB}{0,92,175}
\begin{document}

\title{Learning Adaptive Dynamical Features via Multi-$\tau$ Liquid-Mamba for All-in-one Image Restoration}

\author{Hu Gao, Changshuo Wang, Yulong Chen and Lizhuang Ma$^{\dag}$
        % <-this % stops a space
\thanks{Hu Gao, Changshuo Wang and  Lizhuang Ma are  with the Department of Computer Science, Shanghai Jiao Tong University, Shanghai 200240, China (e-mail: gao\_h@sjtu.edu.cn, 1145209406@sjtu.edu.cn, lzma@sjtu.edu.cn).

Yulong Chen is with the Department of Architecture and Design, Harbin Institute of Technology, Heilongjiang, China (e-mail: llong\_c@hit.edu.cn)
}}



\maketitle

\begin{abstract}
Image restoration aims to recover high-quality images from degraded observations. Recent Mamba-based image restoration models have demonstrated strong potential in modeling long-range dependencies with linear complexity. However, most existing designs still rely on a single state-evolution timescale, which limits their adaptability to spatially heterogeneous and task-dependent degradation patterns in all-in-one image restoration. In this paper, we propose Multi-$\tau$ Liquid-Mamba, an adaptive state space module that introduces input-conditioned multi-timescale liquid discretization into selective state space modeling. Instead of changing the overall selective scan pipeline, the proposed module modulates the effective discretization steps of multiple dynamical branches and adaptively fuses their responses according to degradation-aware gating weights. This design allows the model to capture both fast-varying local details and slowly evolving global structures while preserving the linear scaling property of Mamba with respect to sequence length.
Importantly, Multi-$\tau$ Liquid-Mamba modulates the effective transition dynamics while preserving the original selective parameterization and hardware-efficient selective scan mechanism, making it a plug-and-play module that can be seamlessly integrated into existing Mamba-based architectures.
Built upon this framework, we develop a Multi-$\tau$ Liquid-Mamba Image Restoration Network (MLMIR) for all-in-one image restoration. Extensive experiments on a wide range of restoration benchmarks demonstrate that MLMIR consistently achieves state-of-the-art performance in all-in-one image restoration while remaining highly competitive in task-aligned restoration settings. 

\end{abstract}


\section{Introduction}
Image restoration aims to recover high-quality images from degraded observations. Over the past decades, extensive efforts have been devoted to addressing various degradation types, including image denoising, deblurring, deraining, desnowing, and dehazing. Recently, with the increasing demand for practical applications, the research focus has gradually shifted from task-specific and task-aligned restoration toward more challenging all-in-one image restoration, where a single model is expected to handle multiple degradations simultaneously. Compared with task-aligned settings, all-in-one restoration requires stronger representation capability and greater adaptability to diverse degradation characteristics.

Early image restoration methods mainly relied on handcrafted priors, such as sparsity constraints, total variation regularization, and physical degradation modeling~\cite{2011Single, 10558778, 11342300}. Although these methods exhibit good interpretability, their restoration performance depends heavily on strong assumptions about degradation factors. In real-world scenarios, where degradations are often uncertain or unknown, accurate modeling becomes challenging, leading to limited generalization and unstable performance.
With the rapid development of deep learning, convolutional neural networks (CNNs) have significantly advanced image restoration by learning powerful degradation-specific representations from large-scale datasets~\cite{gao2026emphasizing, starir11429607,gao2025mixed}. Nevertheless, the inherently local receptive field of convolution operations makes it difficult to model long-range dependencies and global contextual information, which are crucial for recovering severely degraded images.
To address this limitation, Transformer-based approaches have been extensively explored in image restoration~\cite{DswinIR11304568, baryir11417902,Allrestor11367271}. By leveraging self-attention mechanisms, Transformers can capture long-range interactions and global image structures. Numerous studies have demonstrated superior restoration performance. However, the quadratic computational complexity of self-attention with respect to image resolution substantially increases memory consumption and computational cost, limiting scalability to high-resolution image restoration scenarios.
More recently, State Space Models (SSMs) have emerged as a promising alternative for efficient sequence modeling~\cite{ALGgao2024learning, TSP-MambaZhou_2025_CVPR, guo2025mambair,guo2025mambairv2, gao2025mbmamba}. In particular, the introduction of Mamba has attracted significant attention due to its ability to capture long-range dependencies with linear computational complexity. Benefiting from the selective state space mechanism and efficient selective scan implementation, Mamba-based architectures have rapidly achieved competitive performance in image restoration.

Despite these advantages, current Mamba-based restoration models still suffer from an inherent limitation. 
Specifically, the state transition process is typically governed by a single discretization mechanism, where hidden-state evolution follows a fixed continuous-time system with only partially input-dependent parameterization. Although the selective mechanism allows adaptive modulation of certain state-space parameters, the underlying temporal dynamics remain restricted to a single timescale. This design is not ideal for image restoration, as degradation patterns often exhibit substantial spatial heterogeneity. For instance, smooth regions generally require stable and gradual state evolution to preserve global structures, whereas edges and textured areas benefit from faster dynamics to recover fine details. Consequently, a single-timescale dynamical system lacks the flexibility needed to effectively accommodate these diverse restoration demands.

More importantly, this limitation restricts the applicability of existing Mamba architectures to all-in-one image restoration. In real-world scenarios, images are often corrupted by multiple degradation types, each characterized by distinct spatial distributions and degradation behaviors. To handle such diversity, an all-in-one restoration model should be able to automatically adapt its feature evolution process according to local image content without requiring prior knowledge of the degradation type. However, existing Mamba-based methods largely rely on a unified state evolution mechanism across all image regions, making it difficult to adjust restoration dynamics to different degradation patterns. As a result, the learned representations may become biased toward specific degradations, thereby limiting both generalization and adaptive restoration performance in complex degradation environments. From a dynamical systems perspective, robust all-in-one restoration requires a model capable of capturing feature evolution at multiple timescales while adaptively selecting appropriate dynamics conditioned on the input. Unfortunately, such flexibility remains largely absent from current Mamba-based restoration frameworks.
\begin{figure} % use float package if you want it here
    \centerline{\includegraphics[width=1\linewidth]{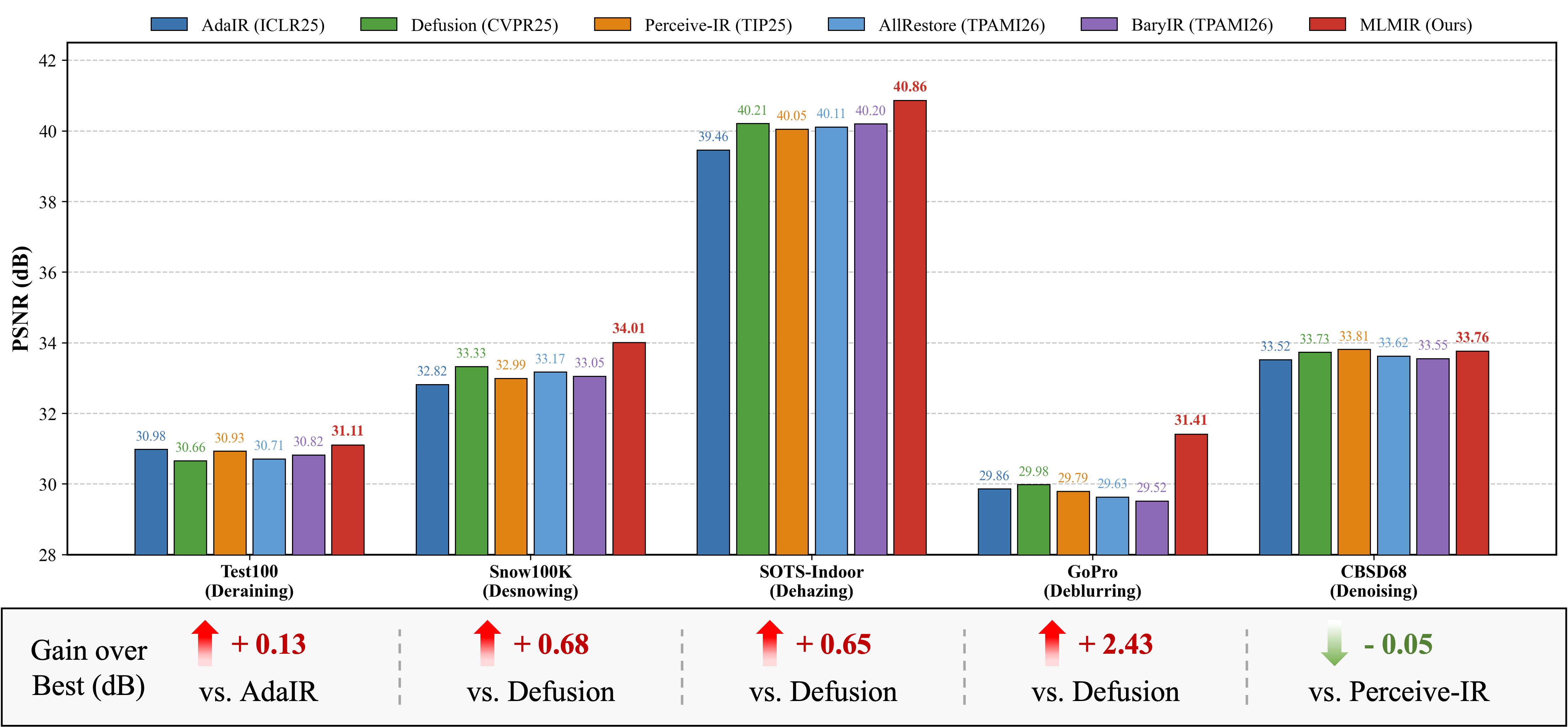}}
	\caption{Comparison of representative all-in-one image restoration methods across five restoration benchmarks. MLMIR achieves the highest overall performance and consistently outperforms the strongest competing baseline on most tasks.}
 \label{fig:res}
\end{figure}

To address these limitations, we propose Multi-$\tau$ Liquid-Mamba, a novel plug-and-play adaptive state space framework for image restoration. Unlike existing Mamba architectures that rely on fixed or only partially input-conditioned state transition dynamics, the proposed framework introduces continuous-time liquid neural dynamics into selective state space modeling. By incorporating multiple adaptive time constants into the discretization process, Multi-$\tau$ Liquid-Mamba enables input-dependent state transitions, allowing the dynamical system to flexibly adjust its temporal evolution according to local degradation characteristics. Specifically, a set of adaptive time constants is estimated from the input features and dynamically fused through a lightweight gating mechanism, resulting in a multi-timescale liquid discretization strategy. This formulation allows hidden-state dynamics to evolve at different rates across image regions, thereby effectively accommodating the substantial spatial variability of real-world degradations. As a consequence, the model can automatically adapt its feature propagation behavior to diverse degradation patterns, significantly enhancing the dynamic expressiveness and adaptive modeling capability of selective state space models.
Importantly, our method modulates the effective transition dynamics while preserving the original selective scan operator and state-space parameterization. Therefore, Multi-$\tau$ Liquid-Mamba can be seamlessly integrated into existing Mamba-based restoration architectures as a plug-and-play module with negligible computational overhead. By enabling content-adaptive multi-timescale state evolution, the proposed framework naturally extends conventional Mamba-based networks from task-specific restoration to the more challenging all-in-one restoration setting.

Building upon the proposed framework, we further develop a Multi-$\tau$ Liquid-Mamba Image Restoration Network (MLMIR) for all-in-one image restoration. MLMIR adopts a hierarchical encoder–decoder architecture and incorporates the proposed Multi-$\tau$ Liquid-Mamba blocks throughout the network to jointly model long-range contextual dependencies and fine-grained local structures. Benefiting from the adaptive multi-timescale dynamics, MLMIR effectively handles diverse degradation patterns.
As shown in Figure~\ref{fig:res}, extensive experiments on multiple all-in-one and task-aligned image restoration benchmarks demonstrate the superiority of the proposed approach.

The main contributions of this work are summarized as follows:
\begin{enumerate}

\item We propose Multi-$\tau$ Liquid-Mamba, a novel adaptive state space framework that extends existing Mamba-based restoration networks from fixed or partially input-conditioned dynamics to fully input-dependent multi-timescale state evolution. As a plug-and-play module, it can be seamlessly integrated into existing architectures, providing a more expressive dynamical formulation for all-in-one image restoration.

\item We design a multi-timescale liquid discretization strategy that jointly models diverse feature evolution rates through learnable liquid time constants and adaptive gating. This mechanism enables content-aware state transitions and significantly enhances the ability of selective state space models to accommodate spatially heterogeneous degradations.

\item We develop MLMIR, a hierarchical image restoration network built upon the proposed framework. Extensive experiments on task-aligned, all-in-one, real-world, unseen-degradation, and mixed-degradation benchmarks demonstrate the effectiveness, generality, and plug-and-play applicability of the proposed module.

\end{enumerate}

\section{Related Works}

\subsection{Image Restoration}

Image restoration (IR) aims to recover high-quality images from degraded observations and represents a fundamental yet highly ill-posed problem in low-level vision. Traditional restoration methods mainly rely on handcrafted priors~\cite{2011Single, 10558778, 11342300}, such as sparsity constraints, low-rank representations, and physical degradation modeling, to regularize the solution space. Although these methods exhibit strong interpretability, their performance is often limited by manually designed assumptions and poor generalization across diverse degradation scenarios.

With the rapid advancement of deep learning, data-driven restoration models have significantly improved restoration quality by learning degradation-aware representations from large-scale datasets~\cite{LSSRgao2024learning,FSNet,starir11429607,CAPTNet10526271,guo2025mambairv2,FDTANetgao2025frequency}. Existing methods can be broadly categorized into task-specific, task-aligned, and all-in-one image restoration according to the number of degradations handled by a single model.

\subsubsection{Task-specific Image Restoration}

Task-specific methods are designed for a particular degradation type, such as image deblurring, denoising, deraining, dehazing, or desnowing. Recent approaches have achieved remarkable progress by exploiting degradation-specific priors and network architectures. For example, ALGNet~\cite{ALGgao2024learning} introduces adaptive local-global feature extraction for motion deblurring, while XYScanNet~\cite{liu2024xyscannet} employs alternating scanning strategies to model long-range spatial dependencies. EfDeRain+~\cite{efderainguo2025efficientderain+} formulates image deraining as a predictive filtering process, avoiding explicit rain-layer modeling. PGH$^2$Net~\cite{PGH2Netisu2025prior} combines bright/dark channel priors with histogram equalization for hierarchical dehazing. Diffusion-based methods, such as UPID-EDM~\cite{upid10.1145/3664647.3680560} and Diff-Unmix~\cite{diffunmix10656884}, further improve restoration quality by leveraging generative priors and iterative denoising processes. Despite their strong performance, task-specific methods generally require separate models for different degradations and thus exhibit limited applicability in real-world scenarios.

\subsubsection{Task-aligned Image Restoration}
To improve generalization across multiple restoration tasks, task-aligned approaches aim to learn a general network capable of handling multiple degradations by sequential training across diverse datasets~\cite{FSNet,gao2025mixed}. Representative methods include SFNet~\cite{SFNet}, FSNet~\cite{FSNet}, ACL~\cite{aclgu2025acl}, StarIR~\cite{starir11429607}, and MHNet~\cite{gao2025mixed}, which exploit frequency-domain modeling, linear attention mechanisms, or hierarchical feature representations to enhance restoration capability. Although these methods improve robustness across different degradations, they still assume a known degradation type during inference and therefore remain limited when confronted with unknown real-world degradations.

\subsubsection{All-in-one Image Restoration}
All-in-one image restoration further extends this paradigm by training a single model to handle multiple degradation types simultaneously~\cite{CAPTNet10526271,baryir11417902,Allrestor11367271}. PromptIR~\cite{potlapalli2023promptir} introduces degradation-aware prompts to guide restoration, while CAPTNet~\cite{CAPTNet10526271} leverages prompt learning to model degradation-specific components. NDR~\cite{NDR10680296} employs neural degradation representations to capture shared degradation characteristics across tasks. BaryIR~\cite{baryir11417902} learns a degradation-invariant representation by minimizing the average Wasserstein distance across multiple degraded distributions, thereby improving generalization to diverse restoration scenarios. 
Recently, large-model-based approaches have shown promising results. VLU-Net~\cite{VLUNetZeng_2025_CVPR} incorporates vision-language representations to identify degradation-aware cues, AutoDIR~\cite{autodir10.1007/978-3-031-73661-2_19} combines vision-language guidance with latent diffusion priors, and Defusion~\cite{DefusionLuo_2025_CVPR} utilizes degradation instruction diffusion for generalized restoration. AdaIR~\cite{cui2025adair} disentangles degradation and content information in both spatial and frequency domains to improve restoration robustness. Perceive-IR~\cite{Perceive-IR10990319} introduces a two-stage design that identifies both degradation types and fine-grained severity levels, achieving strong transferability across restoration tasks. AllRestore~\cite{Allrestor11367271} leverages a composite scene embedding that combines image and text features to characterize degradation conditions, allowing a single model to adaptively handle various restoration tasks.

Despite their strong performance, most existing all-in-one restoration methods rely on prompts, degradation classifiers, diffusion processes, or large vision-language models, leading to substantial computational overhead and increased architectural complexity. In contrast, our method enhances restoration capability from a dynamical systems perspective by introducing adaptive multi-timescale state transitions, without requiring additional degradation recognition modules or external priors.

\subsection{State Space Models for Image Restoration}

State Space Models (SSMs) have recently emerged as a powerful alternative to convolutional and Transformer architectures for long-range dependency modeling. Early works such as S4~\cite{s4gu2021combining} and S5~\cite{s5smith2022simplified} demonstrated that structured state space representations can effectively capture long-range dependencies while maintaining linear computational complexity.

Building upon these advances, Mamba~\cite{gu2023mamba} introduces input-selective state transitions and a hardware-aware selective scan algorithm, significantly improving modeling efficiency and scalability. Owing to its favorable balance between performance and efficiency, Mamba has rapidly attracted attention in image restoration~\cite{gao2025mbmamba,TSP-MambaZhou_2025_CVPR,guo2025mambairv2,guo2025mambair,ALGgao2024learning}. Representative methods such as MambaIR~\cite{guo2025mambair} and MambaIRv2~\cite{guo2025mambairv2} introduce multidirectional scanning and non-causal state modeling to enhance spatial representation learning and contextual aggregation. Other studies explore alternative scanning strategies or hybrid architectures to compensate for the limitations of sequential state evolution. For instance, XYScanNet~\cite{liu2024xyscannet} employs alternating scanning patterns to capture spatial dependencies, while hybrid CNN-Mamba architectures~\cite{ALGgao2024learning} combine local convolutional priors with global state-space modeling.

Despite their success, most existing Mamba-based restoration networks are primarily designed for task-specific or task-aligned restoration settings. 
Several recent studies~\cite{m2resto11284756,AIMVR11209023} have attempted to improve restoration generalization through degradation-aware prompts, auxiliary degradation estimators, or large-scale vision-language priors. However, these approaches usually introduce additional network branches, external guidance modules, or substantial computational overhead. In contrast, we revisit image restoration from a dynamical systems perspective and focus on enhancing the adaptability of the state transition process itself. In this paper, we propose a novel Multi-$\tau$ Liquid-Mamba all-in-one image restoration framework, which introduces adaptive multi-timescale liquid dynamics into selective state space modeling. By learning multiple input-dependent time constants and integrating them into the discretization process, the proposed method significantly improves the dynamic expressiveness of Mamba and enables more flexible state evolution under diverse degradation conditions. Importantly, Multi-$\tau$ Liquid-Mamba only modifies the state transition dynamics while preserving the original selective scan implementation and linear computational complexity. Therefore, it can be seamlessly integrated into existing Mamba-based image restoration architectures as a plug-and-play module, providing a simple yet effective solution for enhancing both task-specific and all-in-one image restoration performance.

\section{Method}
\label{sec:method}
In this section, we first review the necessary preliminaries, then introduce the formulation of Multi-$\tau$ Liquid-Mamba, and finally present the proposed image restoration network built upon it.

\subsection{Preliminaries}

State Space Models (SSMs)~\cite{s4gu2021combining,s5smith2022simplified} provide an efficient framework for modeling long-range dependencies through continuous-time dynamical systems. A linear continuous-time SSM can be formulated as:

\begin{equation}
\frac{d\mathbf{x}(t)}{dt}
=
\mathbf{A}\mathbf{x}(t)
+
\mathbf{B}\mathbf{u}(t),
\label{eq:ssm_continuous}
\end{equation}

\begin{equation}
\mathbf{y}(t)
=
\mathbf{C}\mathbf{x}(t),
\label{eq:ssm_output}
\end{equation}
where $\mathbf{x}(t)\in\mathbb{R}^{N}$ denotes the hidden state, $\mathbf{u}(t)\in\mathbb{R}^{D}$ denotes the input signal, and $\mathbf{y}(t)$ represents the output. The exact solution over a time interval $\Delta$ can be expressed as:

\begin{equation}
\mathbf{x}(t+\Delta)
=
e^{\Delta\mathbf{A}}
\mathbf{x}(t)
+
\int_0^\Delta
e^{(\Delta-s)\mathbf{A}}
\mathbf{B}\mathbf{u}(t+s)\,ds.
\label{eq:ssm_solution}
\end{equation}

Assuming the input is piecewise constant within each interval, i.e., $\mathbf{u}(t+s)=\mathbf{u}_k$, and following a zero-order hold discretization, the continuous system can be discretized into the following form:

\begin{equation}
\mathbf{x}_{k+1}
=
\bar{\mathbf{A}}\mathbf{x}_k
+
\bar{\mathbf{B}}\mathbf{u}_k,
\label{eq:ssm_discrete}
\end{equation}

\begin{equation}
\mathbf{y}_k
=
\mathbf{C}\mathbf{x}_k,
\end{equation}

\begin{equation}
\bar{\mathbf{A}}=e^{\Delta\mathbf{A}},\quad
\bar{\mathbf{B}}=(\Delta\mathbf{A})^{-1}\left(e^{\Delta\mathbf{A}}-\mathbf{I}\right)(\Delta\mathbf{B}).
\end{equation}

This formulation serves as the foundation of modern structured state space models. Recent Mamba~\cite{gu2023mamba} architectures extend SSMs via selective state space mechanisms, introducing input-dependent parameterization that enables content-aware information propagation. Instead of relying on fixed input and output projections, selective SSMs dynamically generate:

\begin{equation}
\mathbf{x}_{k+1}
=
\bar{\mathbf{A}}\mathbf{x}_k
+
\mathbf{B}(\mathbf{u}_k)\mathbf{u}_k,
\label{eq:mamba_update}
\end{equation}

\begin{equation}
\mathbf{y}_k
=
\mathbf{C}(\mathbf{u}_k)\mathbf{x}_k,
\label{eq:mamba_output}
\end{equation}
\begin{equation}
\mathbf{B}_k=\mathbf{B}(\mathbf{u}_k), \quad \mathbf{C}_k=\mathbf{C}(\mathbf{u}_k).
\end{equation}

By leveraging the selective scan algorithm, these recurrent dynamics can be computed with linear complexity with respect to sequence length. Despite their remarkable success, most existing Mamba-based restoration networks~\cite{guo2025mambairv2,aclgu2025acl,TSP-MambaZhou_2025_CVPR} are primarily developed for task-specific or task-aligned restoration scenarios. Their performance gains mainly stem from architectural refinements, feature interaction mechanisms, and improved scanning strategies, while the underlying state transition dynamics remain largely unchanged. In particular, existing methods typically employ a single-timescale discretization scheme to govern hidden-state evolution. Although such a formulation is effective for capturing long-range dependencies, it lacks the flexibility required to accommodate the highly heterogeneous degradation patterns encountered in all-in-one image restoration, where multiple degradation types may coexist and exhibit substantially different restoration characteristics. As a result, the representation capability of current Mamba-based models is inherently constrained when attempting to model diverse degradation distributions within a unified framework.

%%%%%%%%%%%%%%%%%%%%%%%%%%%%%%%%%%%%%%%%%%%%%%%%%%%%%%%%%%%%%%%%%%%%%%%%%
\subsection{Multi-$\tau$ Liquid-Mamba}
\label{sec:multi_tau_liquid_mamba}

Although Mamba introduces input-dependent parameterization for $\mathbf B_k$ and $\mathbf C_k$, the hidden-state evolution is still governed by a single transition operator $\mathbf{\Phi}_k=e^{\mathbf A_0 \Delta_k}$,
where $\mathbf A_0$ denotes the shared state matrix and $\Delta_k$ is generated from a single discretization branch.

Such a formulation implicitly assumes that all degradations can be modeled using a common dynamical system. While this design is effective for long-range sequence modeling, it provides limited flexibility for adapting to the diverse degradation characteristics encountered in all-in-one image restoration. Different degradation types often exhibit distinct restoration dynamics and require different state evolution behaviors, rendering a single-timescale transition mechanism insufficient for jointly modeling multiple degradations within a unified architecture.
For example, image denoising mainly relies on short-range information aggregation, while motion deblurring and haze removal require substantially longer memory horizons to capture global degradation structures.
Consequently, employing a single transition process inevitably limits the adaptability of existing Mamba-based restoration networks.

To address this limitation, we revisit selective state space modeling from the perspective of continuous-time dynamical systems and incorporate adaptive liquid dynamics into the discretization process. Rather than directly modifying the state matrix, we regulate the evolution speed of hidden states through input-dependent time constants. This design retains the computational efficiency and parallelizable selective scan implementation of Mamba, while significantly improving the adaptability and expressiveness of the underlying state transition dynamics.

%%%%%%%%%%%%%%%%%%%%%%%%%%%%%%%%%%%%%%%%%%%%%%%%%%%%%%%%%%%%%%%%%%%%%%%%%
\subsubsection{Liquid State Transition Dynamics}

Liquid Time-Constant Networks (LTCs)~\cite{ltchasani2021liquid} introduce adaptive continuous-time dynamics through input-dependent time constants:
\begin{equation}
\frac{d\mathbf x(t)}{dt}
=
-\frac{\mathbf x(t)}
{\tau(\mathbf u)}
+
f(\mathbf u),
\label{eq:ltc}
\end{equation}
where $\tau(\mathbf u)$ controls the evolution speed of the hidden state.
Inspired by LTCs, we introduce liquid dynamics into selective state space models and consider the following input-dependent continuous-time system:

\begin{equation}
\frac{d\mathbf x(t)}{dt}
=
\mathbf A(\mathbf u)
\mathbf x(t)
+
\mathbf B(\mathbf u)
\mathbf u(t).
\label{eq:liquid_ssm}
\end{equation}

Unlike conventional Liquid-S4~\cite{hasani2023liquid}, directly replacing the state transition matrix with a scalar liquid coefficient may destroy the expressive transition structure learned by Mamba. Therefore, we preserve the original state matrix and introduce adaptive perturbations around the original dynamics.

%%%%%%%%%%%%%%%%%%%%%%%%%%%%%%%%%%%%%%%%%%%%%%%%%%%%%%%%%%%%%%%%%%%%%%%%%

\subsubsection{Multi-Timescale Liquid Dynamics}
Although a single adaptive time constant improves flexibility, it remains insufficient for modeling the diverse degradation characteristics encountered in all-in-one image restoration. In practice, smooth regions, edges, textures, and severe degradations often require different state evolution rates. Therefore, we introduce multiple learnable timescales to capture heterogeneous restoration dynamics.
Unlike conventional LTC formulations, directly replacing the transition matrix with a scalar liquid coefficient may weaken the expressive transition structure learned by Mamba. To avoid this issue, we keep the original Mamba transition matrix $A_0$ as a shared base and introduce a lightweight input-conditioned low-rank residual perturbation around it:

\begin{equation}
\frac{d\mathbf x(t)}{dt}
=
\mathbf A_m(\mathbf u)
\mathbf x(t)
+
\mathbf B(\mathbf u)
\mathbf u(t),
\label{eq:gama}
\end{equation}

\begin{equation}
\mathbf{A}_m(\mathbf{u})
=
\mathbf{A}_0
+
\mathbf{\Gamma}_m(\mathbf{u}),
\label{eq:adaptiveA}
\end{equation}

\begin{equation}
\mathbf{\Gamma}_m(\mathbf{u})
=
\mathbf{U}_m
\operatorname{diag}
\big(
\tau_m(\mathbf{u})
\big)
\mathbf{V}_m^{T}.
\label{eq:gamma}
\end{equation}
where $\mathbf{A}_0$ denotes the base transition matrix inherited from Mamba, $\tau_m(\mathbf{u})$ is the $m$-th adaptive liquid timescale, and $\mathbf{U}_m,\mathbf{V}_m$ are learnable low-rank projections. This formulation does not replace the original transition structure with an independently learned full matrix. Instead, it performs a lightweight residual modulation of the effective transition operator, allowing the dynamics to adapt to image content while retaining the original Mamba-style state-space parameterization. Unlike Liquid-S4, which modifies the convolution kernel through bilinear liquid approximations, our formulation directly modulates the state-transition operator while preserving the selective scan mechanism. Here, preserving the selective scan mechanism means that the scan order, recurrent computation form, and hardware-efficient selective scan implementation are kept unchanged. The proposed low-rank perturbation only changes the branch-wise effective transition parameters used inside the state update, rather than introducing new scanning directions, external degradation estimators, or additional attention-style global interactions.

Compared with directly learning an independent transition matrix for each degradation, the low-rank formulation in Eq.~(\ref{eq:gamma}) preserves the efficient structure of Mamba while enabling adaptive deformation of the underlying dynamical system. Given the hidden representation $\mathbf r_k$, the $m$-th timescale is generated as:

\begin{equation}
\tau_m
=
\operatorname{Softplus}
(
W_{\tau_m}\mathbf r_k
+
b_{\tau_m}
),
\label{eq:tau}
\end{equation}

Each branch therefore evolves under both a distinct dynamical system $\mathbf A_m$ and a distinct evolution speed $\tau_m$, resulting in a family of multi-timescale liquid dynamical systems.

%%%%%%%%%%%%%%%%%%%%%%%%%%%%%%%%%%%%%%%%%%%%%%%%%%%%%%%%%%%%%%%%%%%%%%%%%
\subsubsection{Continuous-to-Discrete State Evolution}

Eq.~(\ref{eq:gama}) apply zero-order hold discretization over interval $\Delta_k \tau_m$ gives

\begin{equation}
\mathbf x_{k+1}^{(m)}
=
\mathbf \Phi_m
\mathbf x_k^{(m)}
+
\mathbf \Psi_m
\mathbf u_k,
\label{eq:discrete_m}
\end{equation}

\begin{equation}
\mathbf \Phi_m
=
\exp
\!\left(
\mathbf A_m
\Delta_k
\tau_m
\right),
\label{eq:Phi_m}
\end{equation}

\begin{equation}
\mathbf \Psi_m
=
\mathbf A_m^{-1}
\left(
\mathbf \Phi_m
-
\mathbf I
\right)
\mathbf B(\mathbf u_k).
\label{eq:Psi_m}
\end{equation}

Compared with the original Mamba transition operator $\mathbf \Phi_0 = e^{\mathbf A_0 \Delta_k}$, the proposed formulation simultaneously adapts both the transition dynamics and the evolution timescale.
Using the Fréchet expansion of the matrix exponential:
\begin{equation}
\begin{aligned}
e^{(\mathbf A_0+\mathbf \Gamma_m)\Delta_k\tau_m}
&=
e^{\mathbf A_0\Delta_k\tau_m}
+
L_{\exp}
(
\mathbf A_0\Delta_k\tau_m,
\mathbf \Gamma_m\Delta_k\tau_m
)
\\
&+
\mathcal O
(
\|\mathbf \Gamma_m\Delta_k\tau_m\|^2
),
\end{aligned}
\end{equation}

\begin{equation}
L_{\exp}
=
\int_0^1
e^{(1-s)\mathbf A_0 \Delta_k\tau_m}
\mathbf \Gamma_m \Delta_k\tau_m
e^{s\mathbf A_0 \Delta_k\tau_m}
ds.
\end{equation}

Therefore, Multi-$\tau$ Liquid-Mamba can be interpreted as learning a continuous family of transition operators around the original Mamba dynamics.

\begin{algorithm}[t]
\caption{Comparison of Mamba and Multi-$\tau$ Liquid-Mamba State Evolution}
\label{alg:mamba_vs_liquid}
\begin{algorithmic}[1]

\State \textbf{Input:} u
\State \textbf{Output:} y

\vspace{3mm}
\Statex \textcolor{blue}{\textbf{--- Mamba (Single Transition Dynamics) ---}}

\State $\mathbf B = \mathbf B(\mathbf u)$
\State $\mathbf C = \mathbf C(\mathbf u)$
\State \textcolor{blue}{$\Delta = \tau_\Delta$}
\State \textcolor{blue}{$\mathbf \Phi_k = \exp(\mathbf A_0 \Delta_k)$}
\State \textcolor{blue}{$\mathbf x_{k+1} = \mathbf \Phi_k \mathbf x_k + \mathbf B_k \mathbf u_k$}

\State \textbf{Return:} \textcolor{blue}{$\mathbf y_{k+1} = \mathbf C_k \mathbf x_{k+1}$}

\vspace{3mm}
\Statex \textcolor{red}{\textbf{--- Multi-$\tau$ Liquid-Mamba (Proposed) ---}}

\State $\mathbf B = \mathbf B(\mathbf u)$
\State $\mathbf C = \mathbf C(\mathbf u)$

\State Extract hidden representation $\mathbf r$

\State \textcolor{red}{$\tau_m(\mathbf u) = \text{Softplus}(W_{\tau_m}\mathbf r + b_{\tau_m}), \quad m=1,\dots,M$}
\State \textcolor{red}{$\mathbf A_m(\mathbf u) = \mathbf A_0 + \mathbf U_m \mathrm{diag}(\tau_m(\mathbf u)) \mathbf V_m^T$}

\State \textcolor{red}{$\Delta \tau_m \rightarrow \mathbf \Phi_m = \exp(\mathbf A_m(\mathbf u)\Delta \tau_m)$}

    \State \textcolor{red}{$\mathbf \Psi_m = \mathbf A_m^{-1}(\mathbf \Phi_m - \mathbf I)\mathbf B$}

    \For{$m = 1$ to $M$}
        \State \textcolor{red}{$\mathbf x_{k+1}^{(m)} = \mathbf \Phi_m \mathbf x_k^{(m)} + \mathbf \Psi_m \mathbf u_k$}
    \EndFor

    \State \textcolor{red}{$\alpha_m = \mathrm{Softmax}(W_{\alpha_m}\mathbf r)$}

    \State \textcolor{red}{$\mathbf x_{k+1} = \sum_{m=1}^{M} \alpha_m \mathbf x_{k+1}^{(m)}$}

    \State \textbf{Return:} \textcolor{red}{$\mathbf y_{k+1} = \mathbf C_k \mathbf x_{k+1}$}

\end{algorithmic}
\end{algorithm}

%%%%%%%%%%%%%%%%%%%%%%%%%%%%%%%%%%%%%%%%%%%%%%%%%%%%%%%%%%%%%%%%%%%%%%%%%
\subsubsection{Adaptive Dynamical Fusion}

To adaptively select suitable restoration dynamics, a gating network predicts the fusion coefficients

\begin{equation}
\alpha_m
=
\frac
{
\exp
(
W_{\alpha_m}\mathbf r_k
+
b_{\alpha_m}
)
}
{
\sum_{j=1}^{M}
\exp
(
W_{\alpha_j}\mathbf r_k
+
b_{\alpha_j}
)
}.
\label{eq:alpha}
\end{equation}

The final hidden state is obtained by aggregating all liquid branches:

\begin{equation}
\begin{aligned}
\mathbf x_{k+1}
&=
\sum_{m=1}^{M}
\alpha_m
\mathbf x_{k+1}^{(m)}
\\
&=
\sum_{m=1}^{M}
\alpha_m
\left[
\mathbf \Phi_m
\mathbf x_k^{(m)}
+
\mathbf \Psi_m
\mathbf u_k
\right].
\label{eq:fusion}
\end{aligned}
\end{equation}

%%%%%%%%%%%%%%%%%%%%%%%%%%%%%%%%%%%%%%%%%%%%%%%%%%%%%%%%%%%%%%%%%%%%%%%%%
Eq.~(\ref{eq:fusion}) reveals that the proposed Multi-$\tau$ Liquid-Mamba can be viewed as an adaptive mixture of multiple dynamical systems operating at distinct evolution rates. As illustrated in Algorithm~\ref{alg:mamba_vs_liquid}, in contrast to conventional Mamba, which relies on a single transition operator, the proposed formulation jointly learns adaptive transition dynamics via $\mathbf A_m(\mathbf u)$, adaptive memory horizons through $\tau_m(\mathbf u)$, and degradation-aware dynamical fusion through $\alpha_m$. Consequently, different degradations can induce distinct state-transition behaviors and memory ranges, enabling the model to capture substantially richer dynamics and achieve stronger adaptability in all-in-one image restoration.

Importantly, the proposed framework modulates the effective transition dynamics through lightweight low-rank residuals while preserving the original selective scan formulation and linear scaling with respect to sequence length. As a result, Multi-$\tau$ Liquid-Mamba can be seamlessly integrated into existing Mamba-based restoration architectures as a plug-and-play module, without changing the scan order or selective scan implementation.

\begin{figure} % use float package if you want it here
    \centerline{\includegraphics[width=1\linewidth]{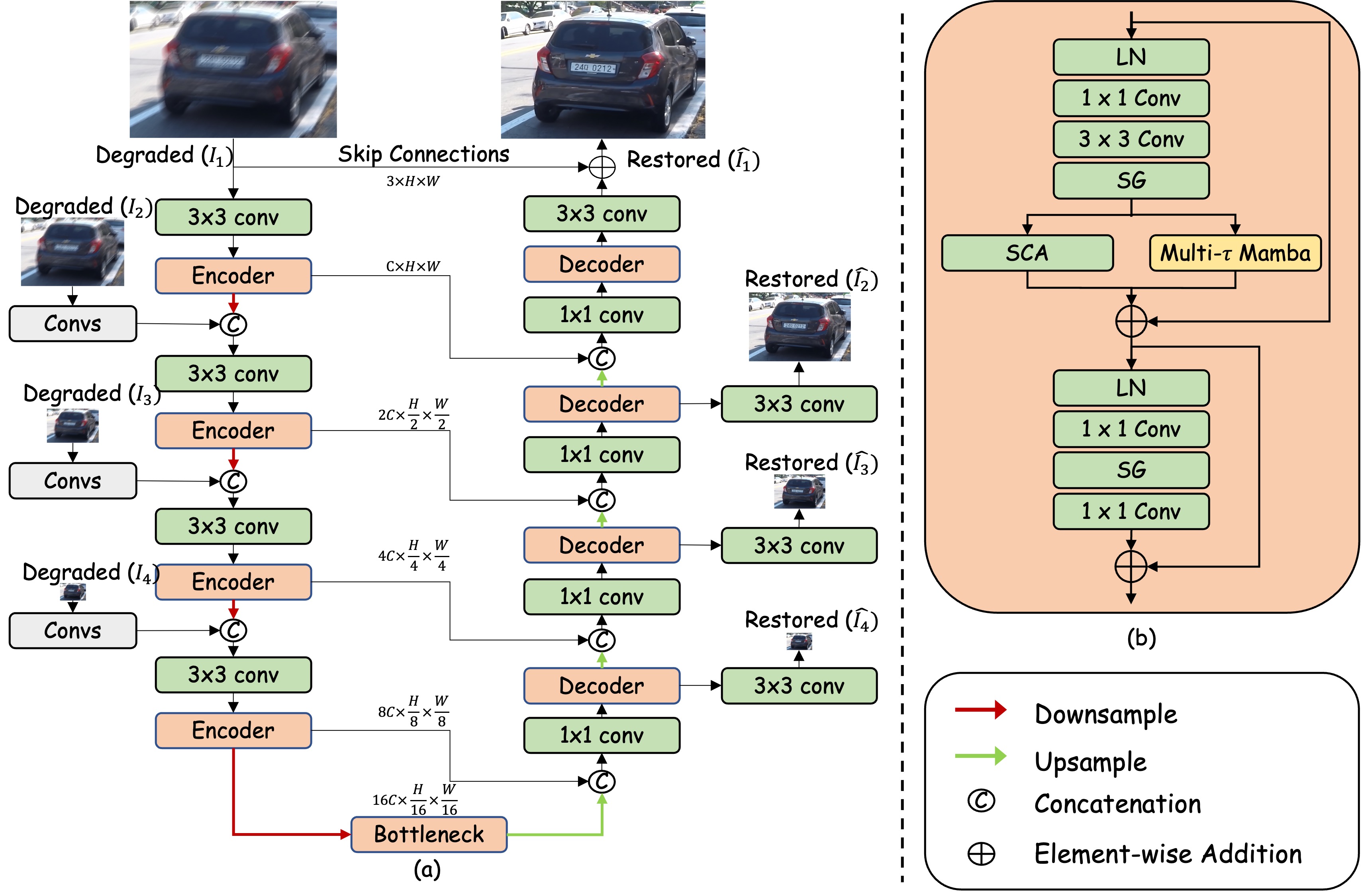}}
	\caption{The overall architecture of MLMIR.}
 \label{fig:network}
\end{figure}

\subsection{Multi-$\tau$ Liquid-Mamba Image Restoration Network}

Built upon the above framework, we propose a Multi-$\tau$ Liquid-Mamba Image Restoration Network (MLMIR) to enhance the adaptability of state space modeling for all-in-one image restoration. The overall framework follows a U-shaped architecture similar to NAFNet~\cite{chen2022simple}, which has demonstrated strong performance and optimization stability in image restoration tasks. Based on this backbone, we replace the original NAFBlock with a Multi-$\tau$ Liquid-Mamba enhanced block, enabling more flexible and input-adaptive state evolution.
As shown in Figure~\ref{fig:network}, the network adopts a multi-scale encoder–decoder paradigm with skip connections to preserve both global context and fine-grained spatial details. Unlike standard designs that rely on single-path feature propagation, our model introduces a multi-input multi-output mechanism across stages, allowing features from different scales to interact more effectively and improving cross-scale information exchange. The key component of the proposed method is the Multi-$\tau$ Liquid-Mamba module, which extends conventional state space models by introducing multiple temporal evolution scales. 

Our model architecture follows a design strategy similar to NAFNet. Specifically, we adopt a block configuration of [1, 1, 1, 28] for each encoder stage, a single block at the bottleneck, and [1, 1, 1, 1] blocks for each decoder stage. This lightweight yet effective design ensures a balanced trade-off between representation capacity and computational efficiency across different resolution levels.
To optimize the network, we employ a composite loss function defined as:
\begin{equation}
\begin{aligned}
\label{eq:loss1}
L &= \sum_{i=1}^{4}[L_{c}(\hat{J_i},\overline I)  + \delta L_{e}(\hat{J_i},\overline I) + \lambda L_{f}(\hat{J_i},\overline I)] 
\\
L_{c} &= \sqrt{||\hat{J_i} -\overline I||^2 + \epsilon^2}
\\
L_{e} &= \sqrt{||\triangle \hat{J_i} - \triangle \overline I||^2 + \epsilon^2}
\\
L_{f} &= ||\mathcal{F}(\hat{J}_i)-\mathcal{F}(\overline I)||_1
\end{aligned}
\end{equation}
where $i$ denotes the index of multi-stage outputs and their corresponding ground-truth supervision, and $\overline{I}$ represents the ground-truth image. $\mathcal{L}_{c}$ is the Charbonnier loss, where $\epsilon$ is set to $0.001$ across all experiments for numerical stability. $\mathcal{L}_{e}$ is an edge-preserving constraint computed in the gradient domain using the Laplacian operator $\triangle$, which encourages the reconstruction of sharp structural details. $\mathcal{L}_{f}$ is a frequency-domain reconstruction loss, where $\mathcal{F}(\cdot)$ denotes the fast Fourier transform to enforce spectral consistency. The weighting coefficients $\lambda$ and $\delta$ are empirically set to $0.1$ and $0.05$, respectively, following prior works~\cite{Zamir2021MPRNet,FSNet}.

\section{Experiments}
\label{sec:exp}
In this section, we first present the experimental setup, followed by qualitative and quantitative comparison results. Finally, ablation studies are conducted to verify the effectiveness of the proposed method.

\subsection{Experimental Setup}
We evaluate our method under both all-in-one and task-aligned settings.

\subsubsection{Datasets}
\textbf{Task-aligned setting.}
\textbf{i)} Image deraining: we train the deraining model on 13,712 paired clean–rain images collected from multiple datasets~\cite{Rain100,Test100,8099669,7780668}. The trained MLMIR is evaluated on several benchmark datasets, including Rain100H~\cite{Rain100}, Rain100L~\cite{Rain100}, Test100~\cite{Test100}, and Test1200~\cite{DIDMDN}.

\textbf{ii)} Image desnowing: we utilize the Snow100K~\cite{desnownet}, SRRS~\cite{JSTASRchen2020jstasr}, and CSD~\cite{HDCW-Netchen2021all} datasets. Following the protocol in~\cite{FSNet}, we randomly sample 2,500 image pairs for training and 2,000 images for evaluation.

\textbf{iii)} Image dehazing: we evaluate MLMIR on the daytime synthetic subsets of RESIDE~\cite{RESIDEli2018benchmarking}, including the Indoor Training Set (ITS), Outdoor Training Set (OTS), and Synthetic Objective Testing Set (SOTS). We train separate models on ITS and OTS, and evaluate them on SOTS-Indoor and SOTS-Outdoor, each containing 500 paired images.

\textbf{iv)} Image deblurring: we train MLMIR on the GoPro dataset~\cite{Gopro}, which contains 2,103 training pairs and 1,111 test pairs.

\textbf{v)} Image denoising: we train a unified MLMIR model to handle multiple noise levels using a composite dataset consisting of 800 images from DIV2K~\cite{DIK}, 2,650 images from Flickr2K~\cite{lim2017enhanced}, 400 images from BSD500~\cite{BSD500}, and 4,744 images from WED~\cite{ma2016waterloo}. Noisy inputs are generated by adding additive white Gaussian noise with a randomly sampled noise level $\sigma \in \{15, 25, 50\}$. Evaluation is performed on CBSD68~\cite{BSD68}, Urban100~\cite{urban100}, and Kodak24~\cite{kodak}.

\textbf{All-in-one setting.}
For the all-in-one scenario, we construct a unified training set by combining multiple restoration datasets, namely ``Rain+Snow+Haze+Blur+Noise''. During testing, we evaluate on Test100~\cite{Test100} for deraining, Snow100K~\cite{desnownet} for desnowing, SOTS-Indoor~\cite{RESIDEli2018benchmarking} for dehazing, GoPro~\cite{Gopro} for deblurring, and CBSD68~\cite{BSD68} for denoising.

\subsubsection{Evaluation Metrics}
We adopt both full-reference and no-reference metrics for evaluation. Full-reference metrics include Peak Signal-to-Noise Ratio (PSNR), Structural Similarity Index (SSIM), and Learned Perceptual Image Patch Similarity (LPIPS)~\cite{54zhang2018unreasonable}. No-reference metrics include the Underwater Colour Image Quality Evaluation (UCIQE)~\cite{55yang2015underwater}, Underwater Image Quality Measure (UIQM)~\cite{56panetta2015human}, Fog Aware Density Evaluator (FADE)~\cite{58choi2015referenceless}, Blind/Referenceless Image Spatial Quality Evaluator (BRISQUE)~\cite{59mittal2011blind}, and Neural Image Assessment (NIMA)~\cite{60talebi2018nima}. Among them, UCIQE and UIQM are designed for underwater image quality assessment, while FADE, BRISQUE, and NIMA are commonly used for real-world dehazing evaluation. For PSNR, SSIM, UCIQE, UIQM, and NIMA, higher values indicate better performance, whereas lower values are preferred for LPIPS, FADE, and BRISQUE. In all tables, the best and second-best results are highlighted in bold and underlined, respectively.

\begin{table*}
\centering
\caption{Quantitative results in the all-in-one setting with task-specific, task-aligned, and all-in-one  IR methods. Denoising results are reported for the noise level 15. }
\label{tb:allrshbn}
    \resizebox{\linewidth}{!}{
\begin{tabular}{c|c|cccccccccc|cc}
    \hline
   \multirow{2}{*}{Type} & \multirow{2}{*}{Methods} & \multicolumn{2}{c}{Deraining} & \multicolumn{2}{c}{Desnowing} & \multicolumn{2}{c}{Dehazing}& \multicolumn{2}{c}{Deblurring}& \multicolumn{2}{c|}{Denoising}  & \multicolumn{2}{c}{Average} 
    \\
   & &PSNR $\uparrow$ &  SSIM $\uparrow$  &PSNR $\uparrow$ &SSIM $\uparrow$ & PSNR $\uparrow$&SSIM $\uparrow$ &PSNR $\uparrow$ & SSIM $\uparrow$&PSNR $\uparrow$ & SSIM $\uparrow$&PSNR $\uparrow$ & SSIM $\uparrow$
    \\
    \hline\hline
   \multirow{4}{*}{\rotatebox{90}{Specific}}
   &MSP-Former~\cite{mspformer10095605} & 28.89 & 0.891 & 32.56 & 0.949 & 36.56 & 0.978 &  27.97 & 0.859 & 32.45 &0.906 & 31.69 & 0.917
     \\
   &EfDeRain+~\cite{efderainguo2025efficientderain+} & 30.89 & \underline{0.909} &  29.42 & 0.925 & 36.39 & 0.979 & 26.48 & 0.804 & 32.67  & 0.906 & 31.17 & 0.905
     \\
     &PGH$^2$Net~\cite{PGH2Netisu2025prior}& 29.52 & 0.896 & 31.88 & 0.932 & 38.82 & 0.985 & 25.96 & 0.789 & 33.05 & 0.943 & 31.84 & 0.909
     \\
     &ALGNet~\cite{ALGgao2024learning}& 28.99 &0.896 & 30.89 & 0.924 & 36.55 & 0.983 & 29.41 & 0.882 & 32.88 & 0.921 & 31.74 & 0.921
     \\
     \hline
    \multirow{4}{*}{\rotatebox{90}{Aligned}}
    &ECFNet~\cite{gao2026emphasizing} & 30.44 & 0.890 & 32.54 & 0.932 & 37.09 & 0.990 & 27.58 & 0.836 & 33.33 & 0.955 & 32.20 & 0.921
    \\
    &MHNet~\cite{gao2025mixed} & 30.26 & 0.888 & 31.94 & 0.941 & 37.44 & 0.989 & 27.41 & 0.830 & 33.25 & 0.934 & 32.06 & 0.916
     \\
     &PPTformer~\cite{pptformerwang2025intra} & 30.52 & 0.906 & 32.58 & 0.931 & 37.96 & 0.990 & 27.89 & 0.845 & 33.51 & 0.949 & 32.50 & 0.925
     \\
    &ACL~\cite{aclgu2025acl} &30.62 & 0.893 & 32.52 & 0.926 & 38.59 & 0.991 & 28.59 & 0.847 & 33.61 & 0.956 & 32.78 & 0.922
     \\
     \hline
      \multirow{7}{*}{\rotatebox{90}{All-in-One}}
      &AdaIR~\cite{cui2025adair}& \underline{30.98} & 0.908 & 32.82 & 0.946 & 39.46 & 0.995 & 29.86 & \underline{0.890} & 33.52 & 0.959 & 33.33 & \underline{0.940}
      \\
      &VLU-Net~\cite{VLUNetZeng_2025_CVPR}& 30.58 & 0.904 & 33.01 & 0.950 & 39.82 & 0.994 & \underline{30.01} & 0.889 & \underline{33.78} & 0.961 & 33.43 &\underline{0.940}
      \\
      &Perceive-IR~\cite{Perceive-IR10990319}& 30.93 & 0.902 & 32.99 & 0.948 & 40.05 & \textbf{0.996} & 29.79 & 0.889 & \textbf{33.81} & \underline{0.962} & 33.51 & 0.940
        \\
        &Defusion~\cite{DefusionLuo_2025_CVPR}& 30.66 & 0.905 & \underline{33.33} & \underline{0.952} & \underline{40.21} & \underline{0.995}& 29.98 & 0.889 & 33.73& \underline{0.962} & \underline{33.58} & \underline{0.940}
        \\
        &AllRestore~\cite{Allrestor11367271} &30.71 &0.903 & 33.17& 0.951 & 40.11 &0.991 & 29.63 &0.882& 33.62 & 0.959 & 33.45 & 0.937
        \\
        &BaryIR~\cite{baryir11417902} & 30.82 &0.907 & 33.05& 0.948& 40.20& 0.995 & 29.52& 0.881& 33.55 & \textbf{0.963} & 33.43 & 0.939
        \\
      &\cellcolor{gray!20}\textbf{MLMIR(Ours)} &\cellcolor{gray!20}\textbf{31.11} 
      &\cellcolor{gray!20}\textbf{0.918} &\cellcolor{gray!20}\textbf{34.01}  &\cellcolor{gray!20}\textbf{0.961} 
      &\cellcolor{gray!20}\textbf{40.86}
      &\cellcolor{gray!20}\textbf{0.996}
    &\cellcolor{gray!20}\textbf{31.41}
      &\cellcolor{gray!20}\textbf{0.946} 
        &\cellcolor{gray!20}33.76
      &\cellcolor{gray!20}\textbf{0.963} 
      &\cellcolor{gray!20}\textbf{34.23} 
      &\cellcolor{gray!20}\textbf{0.957}
    \\
    \hline
\end{tabular}}
\end{table*}

\begin{figure*} % use float package if you want it here
    \centerline{\includegraphics[width=1\linewidth]{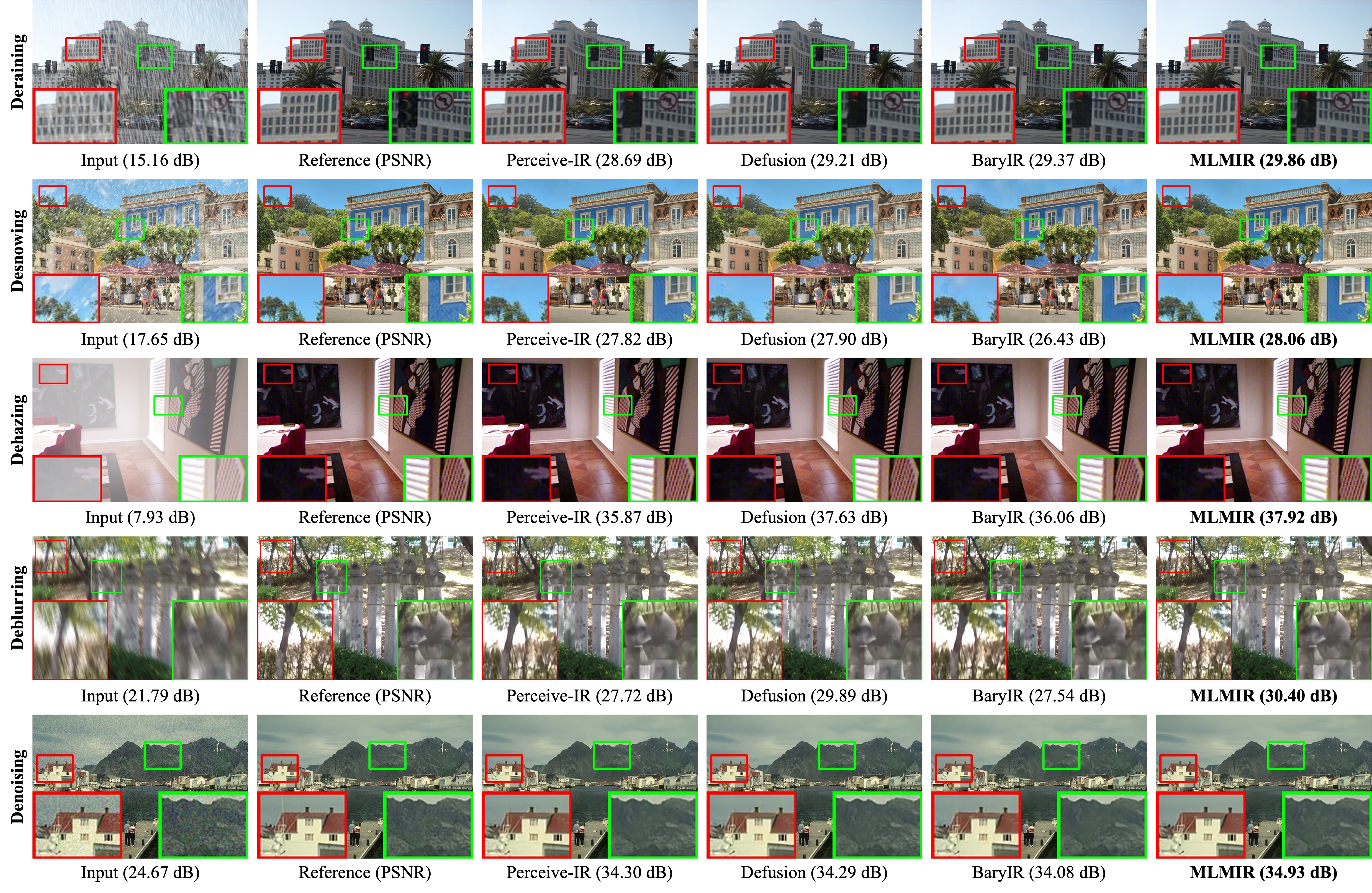}}
	\caption{Qualitative results under the all-in-one experimental setup. Our MLMIR recovers finer details in the reconstructed images.}
 \label{fig:all-in-one}
\end{figure*}

\subsubsection{Training Details}
All models are optimized using the Adam optimizer~\cite{2014Adam} with $\beta_1 = 0.9$ and $\beta_2 = 0.999$. The initial learning rate is set to $2 \times 10^{-4}$ and gradually decayed to $1 \times 10^{-7}$ following a cosine annealing schedule~\cite{2016SGDR}. During training, we randomly crop $256 \times 256$ patches with a batch size of 32, and train for $4 \times 10^5$ iterations. Data augmentation is performed via horizontal and vertical flipping. For fair comparison, all deep learning baselines are retrained or fine-tuned using the hyperparameter settings reported in their original papers.

\subsection{All-in-one Setting Results}
Table~\ref{tb:allrshbn} presents a comprehensive comparison among task-specific, task-aligned, and all-in-one image restoration methods under the challenging R+S+H+B+N setting. Overall, the results clearly demonstrate the superior effectiveness and robustness of the proposed MLMIR across diverse degradation types.

\textbf{Comparison with task-specific methods.}
Although task-specific approaches such as PGH$^2$Net~\cite{PGH2Netisu2025prior} and EfDeRain+~\cite{efderainguo2025efficientderain+} achieve competitive performance on their respective target degradations, their generalization ability across tasks remains limited, leading to suboptimal average performance. In contrast, MLMIR consistently outperforms all task-specific baselines, achieving a clear gain of +2.39 dB over the best task-specific average PSNR (31.84 dB of PGH$^2$Net vs. 34.23 dB of MLMIR). This substantial improvement highlights the advantage of learning unified degradation-aware dynamics rather than task-isolated modeling.

\textbf{Comparison with task-aligned methods.}
Task-aligned methods improve performance by sequentially training across different degradation datasets. However, they still rely on relatively fixed transition mechanisms. As shown in Table~\ref{tb:allrshbn}, MLMIR surpasses the strongest task-aligned baseline ACL~\cite{aclgu2025acl} by +1.45 dB in average PSNR (32.78 dB vs. 34.23 dB). This indicates that explicitly modeling adaptive state-transition dynamics is more effective  in capturing heterogeneous degradation distributions.

\textbf{Comparison with all-in-one methods.}
When compared with recent state-of-the-art all-in-one restoration models, MLMIR still demonstrates consistent and significant improvements. In particular, MLMIR achieves the best overall performance with 34.23 dB PSNR and 0.957 SSIM, outperforming the strongest competing method Defusion~\cite{DefusionLuo_2025_CVPR} by +0.65 dB in average PSNR (33.58 dB $\rightarrow$ 34.23 dB). 

Beyond the average metrics, MLMIR achieves the best or second-best performance across most individual tasks. Specifically, it obtains the highest PSNR on deraining (31.11 dB), desnowing (34.01 dB), and dehazing (40.86 dB), while also delivering competitive performance on deblurring (31.41 dB, best among all methods) and denoising (33.76 dB, close to the best score of 33.81 dB achieved by Perceive-IR~\cite{Perceive-IR10990319}). These consistent gains across different degradation types demonstrate the effectiveness of the proposed multi-timescale liquid dynamics in modeling heterogeneous restoration processes.

Figure~\ref{fig:all-in-one} further provides qualitative comparisons, where MLMIR produces sharper structures and more faithful textures compared with existing methods.

\subsection{Task-aligned Setting Results}
To demonstrate that the proposed MLMIR is effective not only for all-in-one restoration but also for task-specific scenarios, we conduct experiments on five representative image restoration tasks: image deraining, image desnowing, image dehazing, image deblurring and image denoising.

\begin{figure} % use float package if you want it here
    \centerline{\includegraphics[width=1\linewidth]{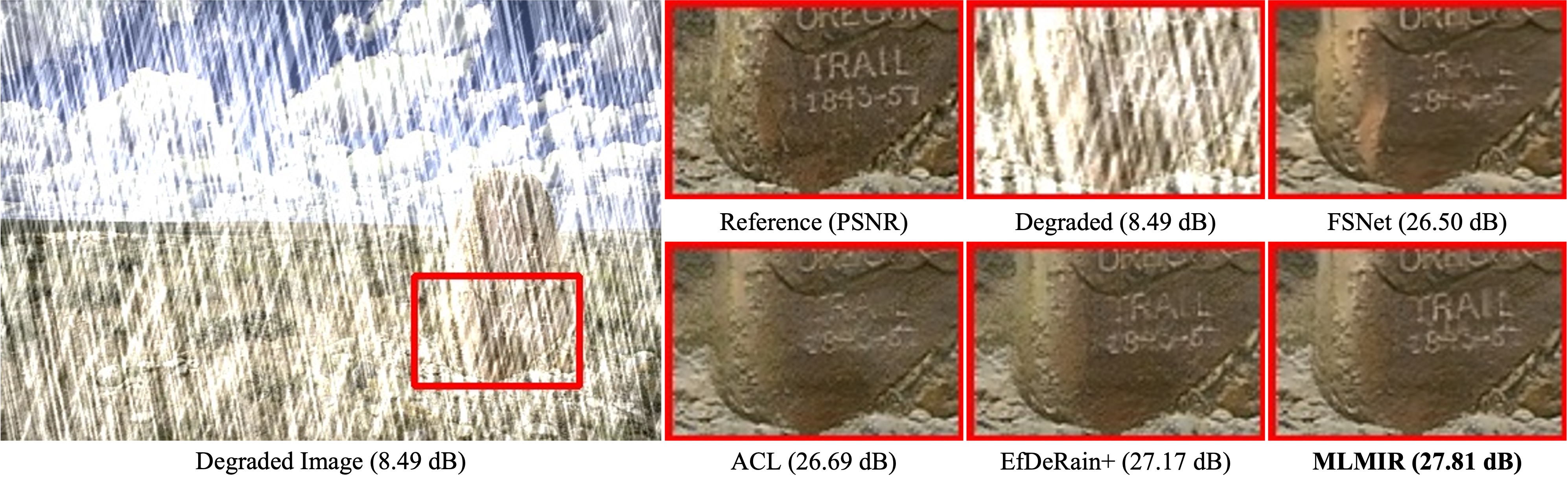}}
	\caption{Image deraining results under the task-aligned experimental setup.}
 \label{fig:rain}
\end{figure}

\begin{table*}
\centering
\caption{Image deraining results in the task-aligned setting.}
\label{tb:derain}
    \resizebox{\linewidth}{!}{
\begin{tabular}{c|cccccccc|cc}
    \hline
    \multicolumn{1}{c|}{} & \multicolumn{2}{c}{Test100}  & \multicolumn{2}{c}{Test1200} & \multicolumn{2}{c}{Rain100H} & \multicolumn{2}{c|}{Rain100L} & \multicolumn{2}{c}{Average} 
    \\
   Methods &PSNR $\uparrow$ &  SSIM $\uparrow$  & PSNR $\uparrow$ & SSIM $\uparrow$ &PSNR $\uparrow$ &SSIM $\uparrow$ & PSNR $\uparrow$&SSIM $\uparrow$ &PSNR $\uparrow$ & SSIM $\uparrow$
    \\
    \hline\hline
    FSNet~\cite{FSNet} &31.05&0.919 &33.08&0.916 & 31.77&0.906 &38.00 & 0.972 &33.48 &0.928
    \\
    MHNet~\cite{gao2025mixed} &31.25 &0.901 &\underline{33.45} &0.925 &31.08 &0.899 &\underline{40.04} &\textbf{0.985} &33.96 &0.928
     \\
     PPTformer~\cite{pptformerwang2025intra} & 31.48 & \underline{0.922} & 33.39 & 0.911 &  31.77 & 0.907
     & 39.33 &\underline{0.983} & 33.99 & 0.931
     \\
 ACL~\cite{aclgu2025acl} &\underline{31.51} & 0.914 & 33.27 &\underline{0.928}  & 32.22 & 0.920 & 39.18 &\underline{0.983}  & 34.05 & 0.936
     \\
     EfDeRain+~\cite{efderainguo2025efficientderain+} &31.10 &0.911 &33.12 & 0.925 & \textbf{34.57} &\textbf{0.957} & 39.03 & 0.972 &\underline{34.46} & \underline{0.941}
     \\
      \hline
    \rowcolor{gray!20}  \textbf{MLMIR(Ours)}  & \textbf{32.66}	&\textbf{0.925}	&\textbf{35.02}	&\textbf{0.941}	&\underline{33.07}	&\underline{0.921}	&\textbf{40.38}	&\textbf{0.985}	&\textbf{35.28}	&\textbf{0.943}
    \\
    \hline
\end{tabular}}
\end{table*}

\subsubsection{Image Deraining} 
Following the evaluation protocol in prior work~\cite{gao2025mixed}, we report PSNR and SSIM on the Y channel in the YCbCr color space for the deraining task. Table~\ref{tb:derain} summarizes the quantitative results under the task-aligned setting.
Overall, the proposed MLMIR consistently achieves superior performance across all benchmark datasets. On Test100~\cite{Test100} and Test1200~\cite{DIDMDN}, MLMIR achieves 32.66 dB and 35.02 dB in PSNR, respectively, outperforming all competing methods by a clear margin. 
For Rain100H~\cite{Rain100}, EfDeRain+~\cite{efderainguo2025efficientderain+} obtains a slightly higher PSNR (34.57 dB), while MLMIR still delivers competitive results (33.07 dB) and maintains the best overall performance across datasets. In particular, MLMIR achieves the best average PSNR (35.28 dB) and SSIM (0.943), surpassing all competing methods, including the strongest baseline EfDeRain+ (34.46 dB PSNR and 0.941 SSIM). This demonstrates that the proposed method achieves a better trade-off between different rain degradation types and provides more consistent restoration performance across diverse scenarios.
Figure~\ref{fig:rain} further shows that MLMIR produces visually cleaner results with significantly reduced color distortion compared with existing methods. In addition, it reconstructs finer textures and sharper structural details in heavily degraded regions, highlighting its effectiveness in restoring both global consistency and local detail fidelity.

\begin{table}
    \centering
        \caption{Image  desnowing results in the task-aligned setting.}
    \label{tab:snow}
    \resizebox{\linewidth}{!}{
    \begin{tabular}{c|cccccc}
    \hline
    \multicolumn{1}{c|}{} & \multicolumn{2}{c}{CSD}  & \multicolumn{2}{c}{SRRS} & \multicolumn{2}{c}{Snow100K}
    \\
   Methods & PSNR $\uparrow$ & SSIM $\uparrow$  & PSNR $\uparrow$ & SSIM $\uparrow$ & PSNR $\uparrow$ & SSIM $\uparrow$
   \\
   \hline
   \hline
 NAFNet~\cite{chen2022simple} &33.13 &0.96 &29.72 &0.94 &32.41 &\underline{0.95}
\\
FocalNet~\cite{focalnetcui2023focal} &37.18 &\textbf{0.99} &31.34 &\textbf{0.98} &33.53 &\underline{0.95}
\\
MSP-Former~\cite{mspformer10095605} & 33.75 &0.96 &30.76 &\underline{0.95} &33.43& \textbf{0.96}
\\
IRNeXt~\cite{IRNeXt} &37.29 &\textbf{0.99} &31.91&\textbf{0.98} &33.61 &\underline{0.95}
\\
PEUNet~\cite{PEUNet10830558}&37.28 &0.97 &31.89 &\textbf{0.98} &34.11 &\textbf{0.96}
\\
PW-FNet~\cite{pwfnet11433521}&-&-&32.74&\textbf{0.98}&\textbf{34.50}&\underline{0.95}
\\
 StarIR~\cite{starir11429607}  & \textbf{38.57} & \textbf{0.99} & \underline{32.75} & \textbf{0.98} & \underline{34.46} & \textbf{0.96}
\\
\hline
\rowcolor{gray!20} \textbf{MLMIR(Ours)} &\underline{38.35}	&\underline{0.98}  &\textbf{33.86}	&\textbf{0.98}		&34.45&\textbf{0.96}

         \\
         \hline
    \end{tabular}}
\end{table}

\subsubsection{Image Desnowing}
\begin{figure} % use float package if you want it here
    \centerline{\includegraphics[width=1\linewidth]{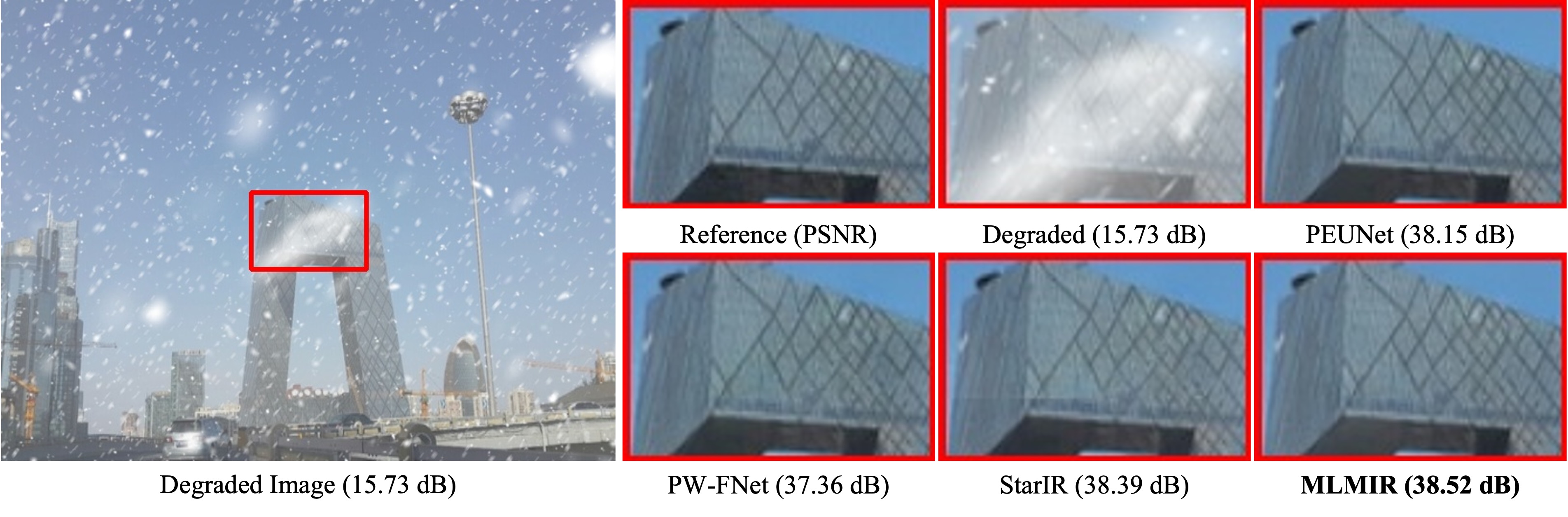}}
	\caption{Image desnowing results under the task-aligned experimental setup.}
 \label{fig:snow}
\end{figure}

Table~\ref{tab:snow} reports quantitative comparisons for image desnowing on three benchmark datasets, including CSD, SRRS, and Snow100K. Overall, the proposed MLMIR achieves highly competitive and consistently strong performance across all datasets.
On the CSD dataset~\cite{HDCW-Netchen2021all}, MLMIR achieves 38.35 dB PSNR and 0.98 SSIM, delivering competitive performance compared with the best baseline StarIR~\cite{starir11429607}, which obtains the highest PSNR (38.57 dB). On SRRS~\cite{JSTASRchen2020jstasr}, MLMIR reaches 33.86 dB PSNR, outperforming all competing methods by a clear margin, including StarIR (32.75 dB) and PW-FNet (32.74 dB). On Snow100K~\cite{desnownet}, MLMIR achieves 34.45 dB PSNR and 0.96 SSIM, matching or exceeding state-of-the-art performance, and remaining highly competitive with the best-performing methods such as PEUNet~\cite{PEUNet10830558} (34.11 dB) and StarIR (34.46 dB).
Figure~\ref{fig:snow} presents qualitative comparisons for image desnowing. As shown, existing methods often suffer from residual snow artifacts or over-smoothing in background regions. In contrast, MLMIR produces cleaner restored images with more faithful structural preservation and sharper texture details, especially in heavily snow-occluded regions.

\begin{figure} % use float package if you want it here
    \centerline{\includegraphics[width=1\linewidth]{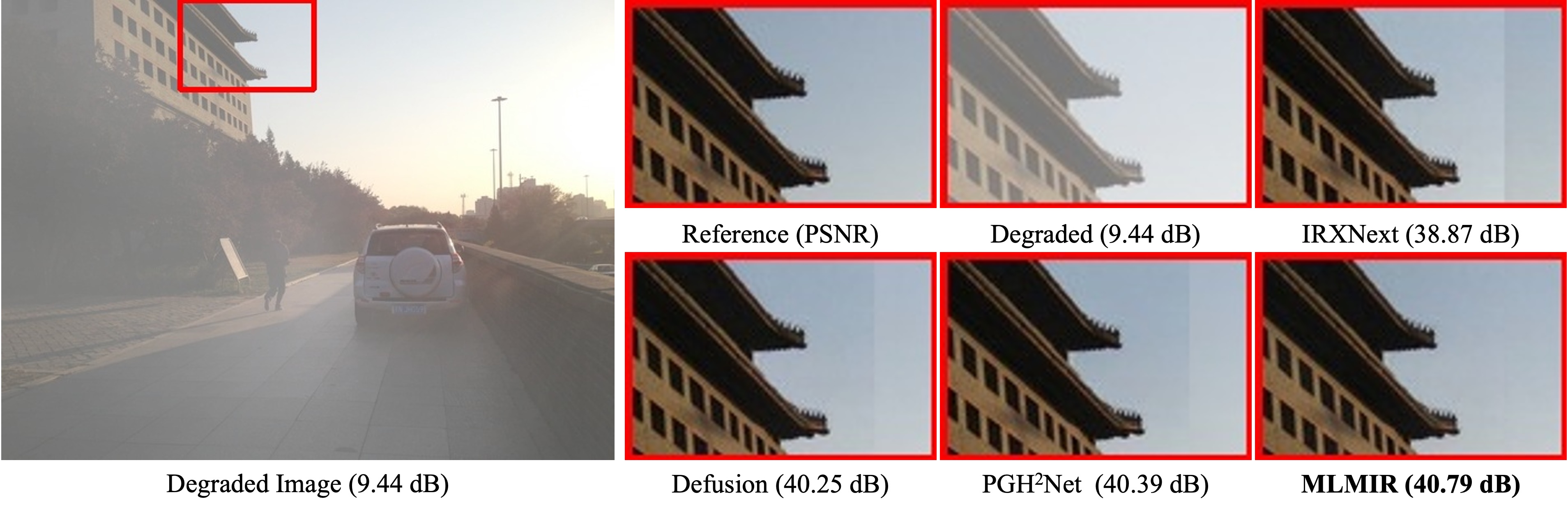}}
	\caption{Image dehazing results under the task-aligned experimental setup.}
 \label{fig:haze}
\end{figure}

\begin{table}
    \centering
        \caption{Image dehazing results in the task-aligned setting.}
    \label{tab:sot}
    \resizebox{\linewidth}{!}{
    \begin{tabular}{c|cccc}
    \hline
    \multicolumn{1}{c|}{} & \multicolumn{2}{c}{SOTS-Indoor}  & \multicolumn{2}{c}{SOTS-Outdoor} 
    \\
   Methods & PSNR $\uparrow$ & SSIM $\uparrow$  & PSNR $\uparrow$ & SSIM $\uparrow$ 
   \\
   \hline
   \hline
IRNext~\cite{IRNeXt}&41.21 &\underline{0.996} &\underline{39.18} &\textbf{0.996}       
\\
DEA-Net-CR~\cite{deanetchen2024dea} & 41.31&0.995 & 36.59 & 0.990 
\\
 Defusion~\cite{DefusionLuo_2025_CVPR} & 41.65&0.995 & 37.41 & \underline{0.993}
\\
PGH$^2$Net~\cite{PGH2Netisu2025prior}&\underline{41.70} &\underline{0.996}&37.52 &0.989
         \\
         \hline
       \rowcolor{gray!20} \textbf{ MLMIR(Ours)} &\textbf{42.14}	&\textbf{0.997}	&\textbf{39.55}	&0.989
         \\
         \hline
    \end{tabular}}
\end{table}

\begin{figure} % use float package if you want it here
    \centerline{\includegraphics[width=1\linewidth]{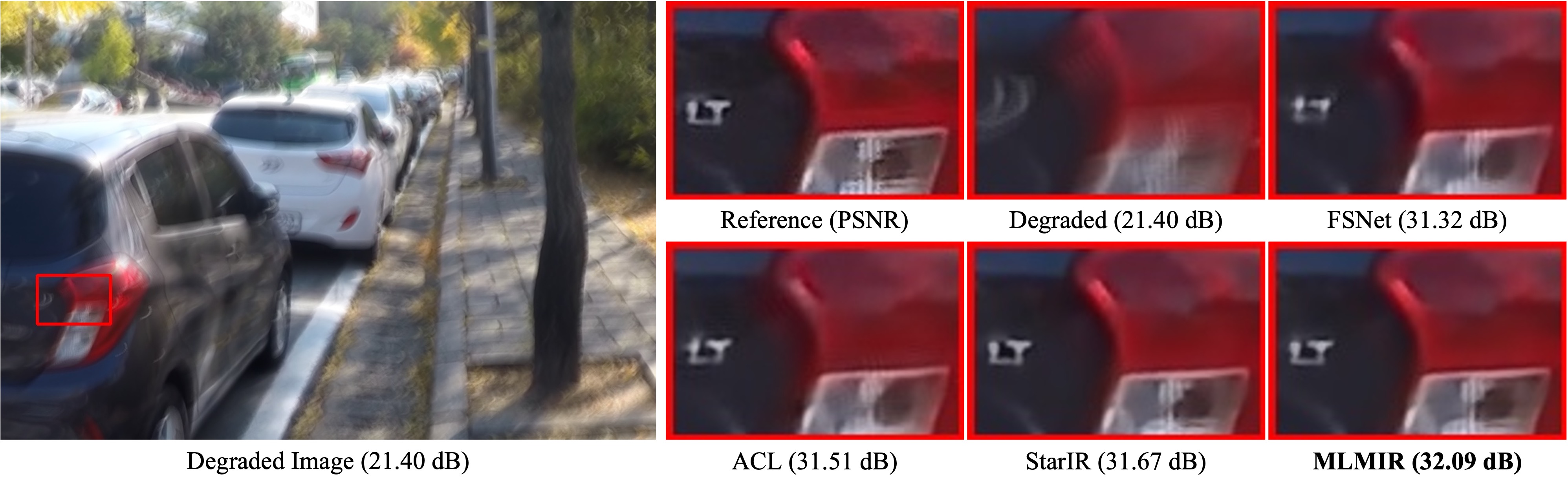}}
	\caption{Image deblurring results under the task-aligned experimental setup.}
 \label{fig:blur}
\end{figure}

\begin{table}
\centering
\caption{Image deblurring results in the task-aligned setting. \label{tb:deblurgh}}

\begin{tabular}{ccccc}
    \hline
    \multicolumn{1}{c}{} & \multicolumn{2}{c}{GoPro}  & \multicolumn{2}{c}{HIDE} 
    \\
   Methods & PSNR $\uparrow$ & SSIM $\uparrow$ & PSNR $\uparrow$ & SSIM $\uparrow$   
    \\
    \hline\hline
    MPRNet~\cite{Zamir2021MPRNet} & 32.66 & 0.959 & 30.96 & 0.939 
    \\
    FSNet~\cite{FSNet} &33.29&0.963 &31.05 & 0.941
    \\
     MambaIR~\cite{guo2025mambair}&33.21 &0.962 &31.01 &0.939
    \\
    ACL~\cite{aclgu2025acl} &33.25 &\underline{0.964} &- &-
    \\
    Omni-Deblurring~\cite{10919160}&33.29 &0.963&31.65&0.947
    \\
    StarIR~\cite{starir11429607} & \underline{34.34}&\textbf{0.970} &\underline{32.12}&\underline{0.951}
    \\
     \hline
       \rowcolor{gray!20} \textbf{ MLMIR(Ours)} &\textbf{34.47}	&\textbf{0.970}	&\textbf{32.92}	&\textbf{0.957}
         \\
    \hline
\end{tabular}
\end{table}

\subsubsection{Image Dehazing}
Table~\ref{tab:sot} reports quantitative comparisons on the SOTS~\cite{RESIDEli2018benchmarking} benchmark for image dehazing. The proposed MLMIR consistently achieves state-of-the-art performance on both SOTS-Indoor and SOTS-Outdoor subsets.
On SOTS-Indoor, MLMIR attains a PSNR of 42.14 dB, outperforming all competing methods. In particular, it surpasses the previous best-performing method PGH$^2$Net~\cite{PGH2Netisu2025prior} (41.70 dB) by +0.44 dB, demonstrating its superior ability to restore clear structures under indoor haze conditions.
On the more challenging SOTS-Outdoor benchmark, MLMIR achieves 39.55 dB PSNR, outperforming the strongest baseline IRNext~\cite{IRNeXt} (39.18 dB) by +0.37 dB. Although DEA-Net-CR~\cite{deanetchen2024dea} and Defusion~\cite{DefusionLuo_2025_CVPR} achieve competitive SSIM values, MLMIR maintains the best overall balance between distortion removal and structural preservation.
Figure~\ref{fig:haze} shows that MLMIR more effectively suppresses haze while preserving fine-grained textures and structural details.

\subsubsection{Image Deblurring}
Table~\ref{tb:deblurgh} reports quantitative comparisons of different image deblurring methods on the GoPro~\cite{Gopro} and HIDE~\cite{HIDE} datasets. Overall, the proposed MLMIR achieves consistently superior performance across both benchmarks.
On the GoPro dataset, MLMIR achieves 34.47 dB PSNR, outperforming all competing methods. In particular, it surpasses the previous state-of-the-art StarIR~\cite{starir11429607} by +0.13 dB, demonstrating its strong capability in modeling complex motion blur degradations. Compared with Omni-Deblurring~\cite{10919160}, MLMIR also shows a clear advantage in PSNR while maintaining comparable structural fidelity.
More importantly, although MLMIR is trained solely on GoPro, it generalizes well to the cross-dataset HIDE benchmark. Specifically, MLMIR achieves 32.92 dB PSNR on HIDE, outperforming the strongest baseline StarIR~\cite{starir11429607} by +0.80 dB in PSNR. Compared with FSNet~\cite{FSNet}, MLMIR further yields a substantial gain of +1.87 dB, highlighting its strong cross-domain generalization ability under severe motion blur conditions.
Figure~\ref{fig:blur} presents qualitative comparisons on representative deblurring cases. As shown,  MLMIR produces sharper structures and more faithful texture restoration, effectively recovering fine details such as edges and repeated patterns.

\begin{figure} % use float package if you want it here
    \centerline{\includegraphics[width=1\linewidth]{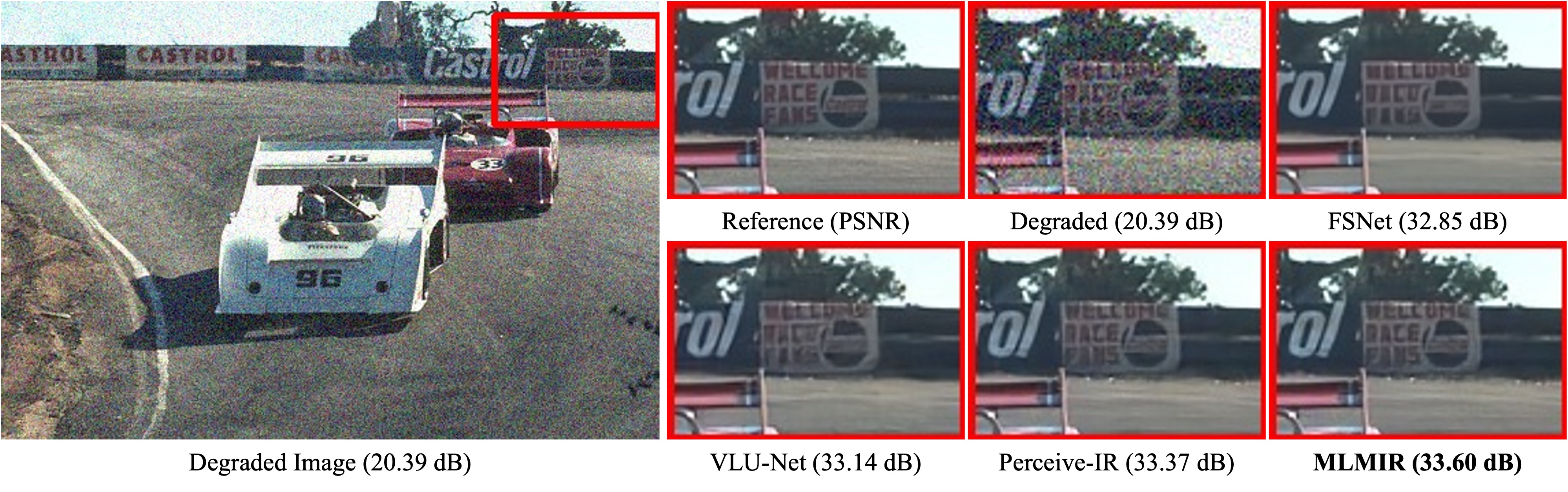}}
	\caption{Image denoising results under the task-aligned experimental setup.}
 \label{fig:noise}
\end{figure}

\begin{table}
    \centering
    \caption{Image denoising results in the task-aligned setting.}
    \label{tab:denoisec}
    \resizebox{\linewidth}{!}{
    \begin{tabular}{c|ccccccccc}
    \hline
      & \multicolumn{3}{c}{CBSD68}  & \multicolumn{3}{c}{Kodak24} & \multicolumn{3}{c}{Urban100} 
    \\
   Methods  & 15  & 25 & 50 & 15  & 25 & 50 & 15  & 25 & 50
   \\
   \hline
   \hline
SFNet~\cite{SFNet} & 34.09 & 31.49 & 28.02 & 34.93 & 32.42 & 29.25&34.19 &32.01 & 29.03
\\
FSNet~\cite{FSNet} & 34.11 & 31.51 & 28.01 & \underline{34.95} & 32.42 & \underline{29.27}&34.15 & 32.04 & 29.15
\\
Perceive-IR~\cite{Perceive-IR10990319} &\underline{34.38} &\textbf{31.74} &\underline{28.53} &34.84 &\underline{32.50} &29.16 &34.86 &32.55 &29.42
\\
VLU-Net~\cite{VLUNetZeng_2025_CVPR}& 34.35 &\underline{31.72} &28.46 &- &- &-&\underline{34.92} &\textbf{32.71} &\textbf{29.61}
\\
\hline
\rowcolor{gray!20} \textbf{MLMIR(Ours)} & \textbf{34.44}	&31.69	&	\textbf{28.50}	&	\textbf{35.38}	&	\textbf{32.89}	&	\textbf{29.86}	&	\textbf{34.94}	&	\underline{32.56}	&	\underline{29.54}

       \\
    \hline
    \end{tabular}}
\end{table}

\begin{table*}
\centering
\caption{Real-world results in the all-in-one  setting with task-specific, task-aligned, and all-in-one  IR methods.}
\label{tb:allrealw}
\begin{tabular}{c|c|cccccccc}
    \hline
   \multirow{2}{*}{Type} & \multirow{2}{*}{Methods} & \multicolumn{3}{c}{RealRain-1k-L} & \multicolumn{3}{c}{RTTS} & \multicolumn{2}{c}{SIDD}
    \\
   & &PSNR $\uparrow$ &  SSIM $\uparrow$ &LPIPS $\downarrow$ &FADE $\downarrow$ &BRISQUE $\downarrow$ &NIMA $\uparrow$&PSNR $\uparrow$ & SSIM $\uparrow$
    \\
    \hline\hline
   \multirow{3}{*}{\rotatebox{90}{Specific}}
   &MSP-Former~\cite{mspformer10095605} &21.09 & 0.744 & 0.428 & 1.721 & 31.039 & 4.322 & 24.12 & 0.457
     \\
   &EfDeRain+~\cite{efderainguo2025efficientderain+} &24.05 & 0.762 & 0.398 & 1.393 & 28.865 & 4.457  & 24.07 & 0.475
     \\
     &PGH$^2$Net~\cite{PGH2Netisu2025prior}&21.65 & 0.747 & 0.419 & 1.352 & 28.501 & \underline{4.625} &24.00 & 0.466
     \\
     \hline
    \multirow{3}{*}{\rotatebox{90}{Aligned}}
    &ECFNet~\cite{gao2026emphasizing} & 23.69 & 0.756 & 0.401 & 1.394 & 29.105 &4.372 & 24.33 & 0.471 
    \\
     &PPTformer~\cite{pptformerwang2025intra} & 23.75 & 0.756 & 0.412 &1.405 &28.301 & 4.320 & 24.37 & 0.469
     \\
    &ACL~\cite{aclgu2025acl} &23.96&0.757&0.408 & 1.360&28.962&4.319 & \underline{24.66} & 0.473
     \\
     \hline
      \multirow{4}{*}{\rotatebox{90}{All-in-One}}
      &VLU-Net~\cite{VLUNetZeng_2025_CVPR}& 27.22 & 0.899 & 0.375 & \underline{1.242} & \textbf{23.490} & 4.564 & 24.31 & 0.471
      \\
      &Perceive-IR~\cite{Perceive-IR10990319}& 27.31 & 0.901  & 0.372 & 1.264 & 23.694 &4.613 & 24.53 & \underline{0.497}
        \\
        &Defusion~\cite{DefusionLuo_2025_CVPR}& \underline{27.37} & \textbf{0.903} & \underline{0.371} & 1.277 & \underline{23.520} & 4.615 & 24.52 & 0.495
        \\
      &\cellcolor{gray!20}\textbf{MLMIR(Ours)} & \cellcolor{gray!20}\textbf{27.54} 
      &\cellcolor{gray!20} \underline{0.902} 
      & \cellcolor{gray!20}\textbf{0.370}
      & \cellcolor{gray!20}\textbf{1.233} & \cellcolor{gray!20}\underline{23.520} & \cellcolor{gray!20} \textbf{4.631} 
      & \cellcolor{gray!20}\textbf{24.69}
      &\cellcolor{gray!20} \textbf{0.499}
    \\
    \hline
\end{tabular}
\end{table*}

\subsubsection{Image Denoising}
Table~\ref{tab:denoisec} reports quantitative results of MLMIR under different noise levels ($\sigma=15, 25, 50$) on CBSD68, Kodak24, and Urban100 benchmarks. Overall, the proposed MLMIR consistently achieves competitive or superior performance across different noise intensities and datasets.
On the CBSD68 dataset, MLMIR achieves strong and stable performance across all noise levels, with particularly competitive results at $\sigma=15$ and $\sigma=50$, where it reaches 34.44 dB and 28.50 dB, respectively. At the medium noise level ($\sigma=25$), MLMIR remains comparable to the best-performing methods.
On the Kodak24 dataset, MLMIR achieves 35.38 dB, 32.89 dB, and 29.86 dB for noise levels 15, 25, and 50, respectively, consistently outperforming all competing methods. In particular, under the most challenging setting ($\sigma=50$), MLMIR surpasses FSNet~\cite{FSNet} by +0.59 dB, highlighting its strong ability to handle severe Gaussian noise.
On the Urban100 dataset, which contains rich structural textures, MLMIR achieves 34.94 dB and 29.54 dB at noise levels 15 and 50, respectively, delivering consistently strong performance. Although VLU-Net~\cite{VLUNetZeng_2025_CVPR} achieves slightly higher results at $\sigma=25$, MLMIR remains competitive across all settings and maintains balanced performance across datasets.
Figure~\ref{fig:noise} presents qualitative comparisons of image denoising results. As shown,  MLMIR produces cleaner images with better edge preservation and more faithful texture reconstruction, especially in highly detailed regions.

\subsection{Generalization}
\subsubsection{Zero-shot generalization in real-world scenes}
We further evaluate the zero-shot generalization capability of different methods under the unified training setting R+S+H+B+N on real-world benchmarks, including RealRain-1k-L~\cite{realrain1li2022toward}, RTTS~\cite{Rttsli2018benchmarking}, and SIDD~\cite{ssidabdelhamed2018high}. Table~\ref{tb:allrealw} summarizes the quantitative results.
On the RealRain-1k-L dataset, MLMIR achieves 27.54 dB PSNR and 0.370 LPIPS, clearly outperforming all task-specific and task-aligned methods. Compared with the strongest baseline Defusion~\cite{DefusionLuo_2025_CVPR}, MLMIR improves PSNR by +0.17 dB while achieving slightly better perceptual quality, demonstrating its superior restoration capability under real-world rain degradations.
On the RTTS hazy scene benchmark, MLMIR achieves the best overall performance with a FADE score of 1.233 and the highest NIMA score of 4.631, while also obtaining competitive BRISQUE results. In particular, it outperforms all existing methods in NIMA, indicating better perceptual quality and more natural visual appearance. These results suggest that MLMIR effectively enhances scene visibility while maintaining realistic image statistics under challenging haze conditions.
On the SIDD denoising benchmark, MLMIR achieves 24.69 dB PSNR and 0.499 SSIM, outperforming all competing methods. Compared with the strongest baseline ACL~\cite{aclgu2025acl} (24.66 dB PSNR), MLMIR achieves consistent improvements while also matching or surpassing all all-in-one methods in SSIM, demonstrating strong robustness under real-world sensor noise.
These results demonstrate that the proposed method generalizes effectively to unseen real-world degradations, benefiting from its adaptive multi-timescale state dynamics, which enables robust modeling of complex and mixed corruption patterns in zero-shot scenarios.

\begin{table}
\centering
\caption{Results on the training-unseen degradation type in the all-in-one  setting.}
\label{tb:allundt}
    \resizebox{\linewidth}{!}{
\begin{tabular}{c|ccccc}
    \hline
    \multirow{2}{*}{Methods} & \multicolumn{3}{c}{UIEB} & \multicolumn{2}{c}{C60}
    \\
   & PSNR $\uparrow$ &  SSIM $\uparrow$ &LPIPS $\downarrow$ &UCIQE$\uparrow$ &UIQM$\uparrow$
    \\
    \hline\hline
  EfDeRain+~\cite{efderainguo2025efficientderain+} & 17.66 & 0.741 & 0.377 & 0.480 & 1.866
     \\
ECFNet~\cite{gao2026emphasizing} & 20.49 & 0.856 & 0.223 & 0.539 & 2.517 
    \\
ACL~\cite{aclgu2025acl} & 20.94 & 0.867 & 0.201 & 0.552 & 2.485 
     \\
     VLU-Net~\cite{VLUNetZeng_2025_CVPR}& 21.58 & 0.889 & 0.172 & \underline{0.556} & 2.493
      \\
      Perceive-IR~\cite{Perceive-IR10990319}& \underline{21.77} &\underline{0.891} & \underline{0.169} & \textbf{0.557} & \underline{2.555 }
        \\
      \rowcolor{gray!20}\textbf{MLMIR(Ours)} & \textbf{21.82} & \textbf{0.893} & \textbf{0.161} & \textbf{0.557} & \textbf{2.557}
    \\
    \hline
\end{tabular}}
\end{table}

\subsubsection{Generalization to unknown degradation type}
To further evaluate the generalization ability of different methods, we test models trained under the unified R+S+H+B+N setting on unseen underwater image restoration benchmarks, including UIEB and C60. Table~\ref{tb:allundt} summarizes the quantitative results.
Overall, the proposed MLMIR consistently achieves the best or highly competitive performance across all evaluation metrics, demonstrating strong robustness to previously unseen degradation types.
On the UIEB dataset, MLMIR achieves the best overall performance. Compared with the strongest baseline Perceive-IR~\cite{Perceive-IR10990319}, MLMIR improves PSNR by +0.05 dB and further reduces LPIPS, indicating better perceptual quality and more faithful structural restoration under complex underwater distortions.
On the C60 benchmark, MLMIR achieves 0.557 UCIQE and 2.557 UIQM, matching or slightly surpassing state-of-the-art methods. In particular, MLMIR reaches the best UIQM score, demonstrating superior color correction and perceptual enhancement capability in challenging underwater environments.
Compared with existing all-in-one methods, MLMIR consistently maintains stronger or comparable performance across both reference-based and no-reference metrics.

\begin{table}
\centering
\caption{Results on the training-unseen degradation severity in the all-in-one  setting .}
\label{tb:allunds}
\begin{tabular}{c|cccc}
    \hline
    \multirow{2}{*}{Methods} & \multicolumn{2}{c}{CBSD68} & \multicolumn{2}{c}{Urban100}
    \\
   & 60 & 100 & 60 & 100
    \\
    \hline\hline
     VLU-Net~\cite{VLUNetZeng_2025_CVPR}&  27.11 & 20.59 & 27.64 & 21.53
      \\
      Perceive-IR~\cite{Perceive-IR10990319}& \underline{27.13} & 20.65 & 27.65 & \underline{21.55}
        \\
        Defusion~\cite{DefusionLuo_2025_CVPR}& 27.11 & \textbf{20.72} & \textbf{27.67} & 21.49
        \\
      \rowcolor{gray!20}\textbf{MLMIR(Ours)} & \textbf{27.17} & \underline{20.71} & \underline{27.66} & \textbf{21.58}
    \\
    \hline
\end{tabular}
\end{table}

\subsubsection{Generalization to unknown degradation severity}
To evaluate the robustness of the proposed method under unseen degradation intensities, we train all models using Gaussian noise levels within the range $\sigma \in [15, 25, 50]$, and further test them on substantially higher, unseen noise levels $\sigma=60$ and $\sigma=100$. The corresponding results are reported in Table~\ref{tb:allunds}.
Overall, MLMIR consistently achieves competitive or superior performance across both CBSD68 and Urban100 benchmarks under unseen noise severities, demonstrating strong robustness to distribution shifts in degradation intensity.
On the CBSD68 dataset, MLMIR achieves the best performance at $\sigma=60$ with 27.17 dB PSNR, outperforming all competing methods. At the more challenging noise level $\sigma=100$, MLMIR remains highly competitive with 20.71 dB PSNR, closely matching the best-performing method Defusion~\cite{DefusionLuo_2025_CVPR}. This indicates that the proposed model maintains stable restoration capability even under extremely strong noise corruption.
On the Urban100 dataset, which contains rich structural textures, MLMIR achieves the best performance at $\sigma=100$ with 21.58 dB PSNR, surpassing all competing methods. At $\sigma=60$, MLMIR also maintains competitive results (27.66 dB).

\begin{figure} % use float package if you want it here
    \centerline{\includegraphics[width=1\linewidth]{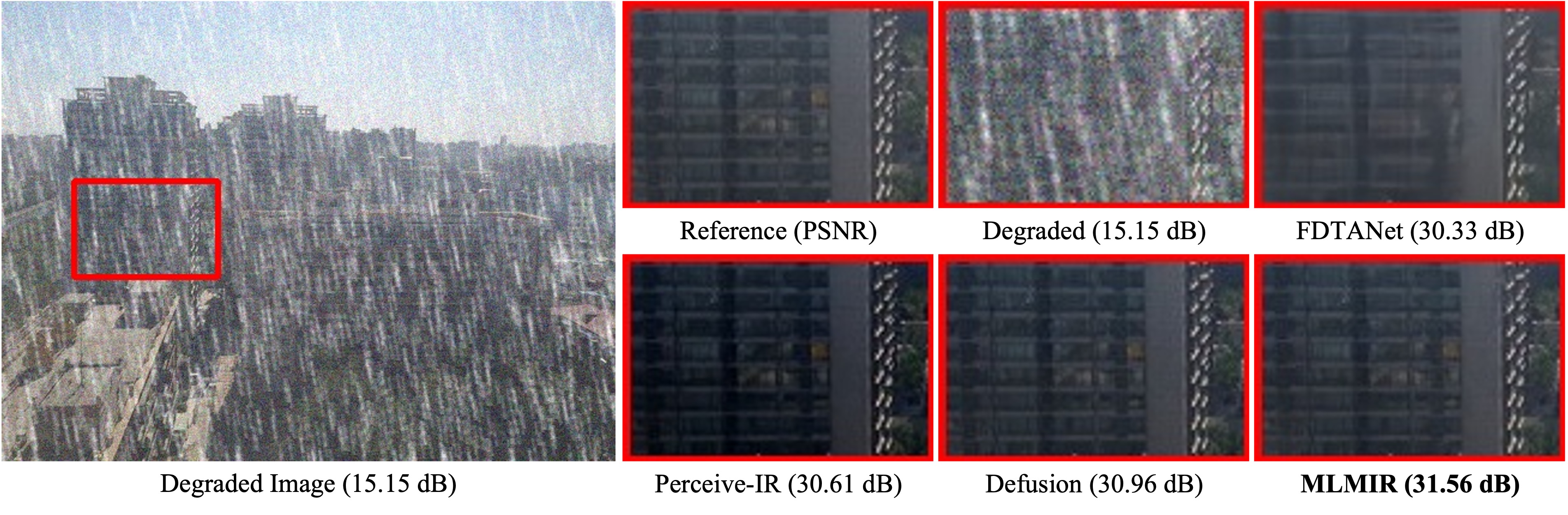}}
	\caption{Multi-degradation (H+R+N) image restoration results under the all-in-one experimental setup.}
 \label{fig:mix}
\end{figure}

\begin{table*}
\centering
\caption{Quantitative results of different models in the all-in-one setting conducted on various combinations of degradation types.}
\label{tb:mix}
\resizebox{\linewidth}{!}{
\begin{tabular}{cccccccc||c}
    \hline
   Methods& H + R + N  & H + R &H + N& R + N & H &R&N&Average
    \\
    \hline\hline
    % Restormer~\cite{Zamir2021Restormer} &23.52/0.792 &25.45/0.836 &26.73/0.863	&25.31/0.815	&29.82/0.931	&28.05/0.873	&27.44/0.837 &26.62/0.850
    %  \\
    % AirNet~\cite{all_conli2022all} &27.41/0.812 &28.28/0.889 &27.59/0.861 &26.98/0.821 &30.22/0.959 &28.27/0.881 & 27.25/0.847 &28.00/0.867
    %    \\
    % U$^2$Former~\cite{u2former} &25.07/0.803 &26.23/0.856 &26.79/0.864	&26.02/0.816	&29.95/0.933	&28.50/0.876	&27.12/0.831 &27.09/0.854
    %  \\
    % PromptIR~\cite{potlapalli2023promptir} &27.54/0.819	&28.43/0.901 &27.99/0.871	&27.05/0.822	&30.46/0.956	&28.78/0.885	&27.92/0.851 &28.31/0.872
    %    \\
    VLU-Net~\cite{VLUNetZeng_2025_CVPR} &29.35/0.842	&30.38/0.935 &\underline{28.79}/0.869	&28.21/0.843	&31.37/0.959	&33.82/0.968	&27.88/0.853 &29.97/0.896
       \\
FDTANet~\cite{FDTANetgao2025frequency}  &29.68/0.846	&30.91/0.937	&\textbf{29.06}/\underline{0.874}	&27.63/0.827	&\textbf{31.91/0.981} &29.13/0.892	&28.88/0.857 &29.60/0.888
\\
Ref-IRT~\cite{REF-IRT}&28.95/0.823	&\underline{32.33/0.961} &28.62/0.872	&31.22/0.831	&30.69/0.961	&31.89/0.902	&29.39/0.899 &30.44/0.893
\\
Perceive-IR~\cite{Perceive-IR10990319}&\underline{30.22}/0.894	&31.55/0.945	&28.32/0.871	&30.52/0.849	&\underline{31.90/0.980}	&38.95/0.983	&31.88/0.923	&31.90/0.921
\\
Defusion~\cite{DefusionLuo_2025_CVPR}&29.66/0.849	&29.93/0.937	&28.11/0.866	&\underline{32.17}/\textbf{0.920}	&31.29/0.969	&\textbf{40.15}/\underline{0.986}	&\textbf{32.72}/\underline{0.925}	&\underline{32.01}/0.922
\\
DSwinIR~\cite{DswinIR11304568} & 30.09/\underline{0.899} & 31.07/0.960 & 28.45/0.873 & 31.29/0.915 & 31.77/0.973 & 39.09/0.982 & \underline{32.22}/0.923 & 31.99/\underline{0.932}
\\
\hline
 \rowcolor{gray!20}\textbf{MLMIR(Ours)} &\textbf{30.45/0.901}	&\textbf{32.52/0.965}	&28.73/\textbf{0.876}	&\textbf{32.23}/\textbf{0.917}	&31.82/0.979	&\underline{40.11}/\textbf{0.987}	&\textbf{32.72/0.929}	&\textbf{32.65/0.936}

    \\
    \hline
\end{tabular}}
\end{table*}

\begin{figure*} % use float package if you want it here
    \centerline{\includegraphics[width=1\linewidth]{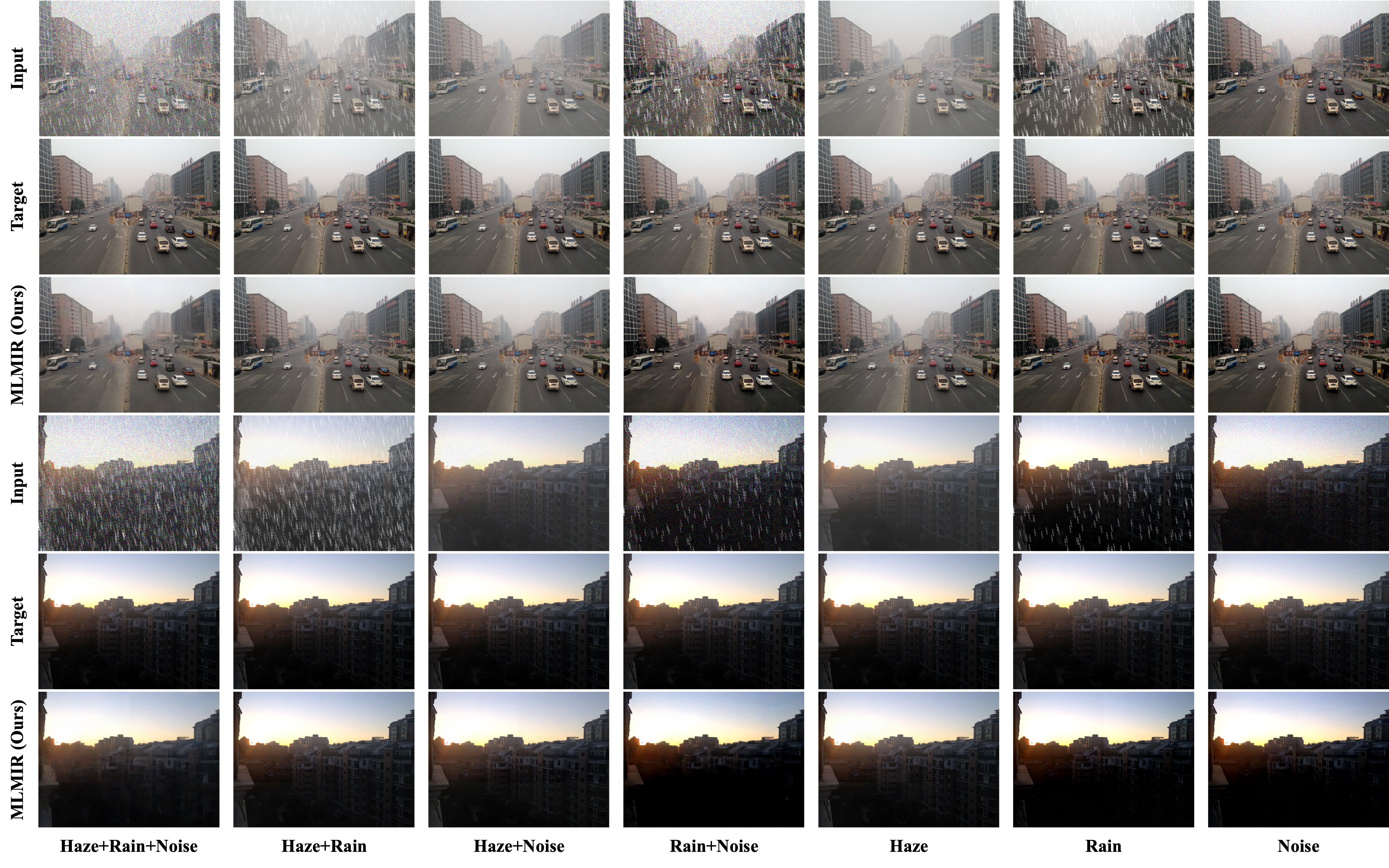}}
	\caption{Multi-degradation image restoration results under the all-in-one setting across various degradation combinations. The top row shows the degraded inputs, the middle row presents the corresponding ground-truth images, and the bottom row displays the restoration results produced by MLMIR.}
 \label{fig:mixm}
\end{figure*}

\subsubsection{Generalization to mixed degradation}
To further evaluate the robustness of different methods under complex real-world conditions, we conduct experiments on mixed degradation scenarios, where multiple corruption types (haze, rain, and noise) are either present individually or combined in various forms. The corresponding quantitative and qualitative results are reported in Table~\ref{tb:mix} and Fig.~\ref{fig:mix}, Fig.~\ref{fig:mixm}, respectively.
MLMIR consistently achieves the best or highly competitive performance across all degradation combinations, demonstrating strong adaptability to both single and compound corruptions.

On mixed degradation settings, MLMIR achieves the best overall performance. In particular, MLMIR significantly improves performance under joint degradation scenarios, where multiple corruption types interact and amplify restoration difficulty. Compared with the strongest baseline Defusion~\cite{DefusionLuo_2025_CVPR}, MLMIR achieves clear gains on both H+R+N and H+R, demonstrating superior capability in handling coupled degradation distributions.
On single-degradation settings, MLMIR maintains consistently strong performance.  These results indicate that the proposed method does not sacrifice single-task performance while improving multi-task generalization.

Figure~\ref{fig:mix} further illustrates qualitative comparisons under multi-degradation scenarios. Existing methods often fail to simultaneously suppress different types of degradations, resulting in residual artifacts, over-smoothing, or color distortion. In contrast, MLMIR produces visually cleaner results with sharper structures and more accurate color recovery, even under severe mixed corruption.
More detailed visualizations in Figure~\ref{fig:mixm} further demonstrate that MLMIR consistently reconstructs fine textures and global structures across different degradation combinations, including H+R+N, H+R, H+N, and R+N. Notably, MLMIR preserves structural consistency close to the ground truth while effectively suppressing compound artifacts, highlighting its strong robustness and generalization capability.

\subsection{Ablation Studies}

\subsubsection{Effect of each component}
We conduct ablation studies under the all-in-one setting to systematically evaluate the contribution of each proposed component. Starting from a baseline model NAFNet~\cite{chen2022simple} without any of the proposed modules (a), the network achieves a PSNR of 32.34 dB.
When the Multi-$\tau$ Liquid-Mamba module is introduced alone (b), the performance is significantly improved to 33.96 dB, yielding a gain of +1.62 dB. This demonstrates that incorporating multi-timescale state modeling effectively enhances the network’s ability to handle complex and spatially varying degradations.
In contrast, introducing only the multi-input and multi-output (MIMO) mechanism (c) leads to a modest improvement to 32.55 dB (+0.21 dB), indicating that cross-scale feature interaction alone provides limited gains when not combined with stronger dynamic modeling. When both the Multi-$\tau$ Liquid-Mamba module and the MIMO design are jointly applied (d), the model achieves the best performance of 34.23 dB, resulting in a substantial improvement of +1.89 dB over the baseline.

\begin{table}
    \centering
    \caption{Ablation study on individual components of MLMIR.}
    \label{tab:abl1}
     \resizebox{\linewidth}{!}{
    \begin{tabular}{ccccc}
    \hline
        Method& Multi-$\tau$ Liquid-Mamba & Multi-in  and Multi-out &  PSNR &$\triangle$ PSNR 
         \\
         \hline
         \hline
         (a) & & & 32.34 & -
         \\
         (b) &\ding{52}& & 33.96 & + 1.62 dB
         \\
          (c) & &\ding{52}  & 32.55 & + 0.21 dB
         \\
        (d)  &\ding{52} &\ding{52}  & 34.23 & + 1.89 dB
         \\
         \hline
    \end{tabular}}
\end{table}

\begin{table}
\centering
\caption{Effect of different combinations of restoration task. }
\label{tb:ablta}
 \resizebox{\linewidth}{!}{
\begin{tabular}{ccccc|ccccc}
    \hline
    \multicolumn{5}{c|}{Task} &  \multirow{2}{*}{Deraining} &\multirow{2}{*}{Desnowing} &\multirow{2}{*}{Dehazing}& \multirow{2}{*}{Deblurring} & \multirow{2}{*}{Denoising}
    \\
    R & S & H & B & N &
    \\
    \hline\hline
    \ding{52}&&&&&32.66 & - &- &- &-
    \\
    &\ding{52}&&&&-&34.45&-&-&-
    \\
    &&\ding{52}&&&-&-&42.14&-&-
    \\
    &&&\ding{52}&&-&-&-&34.47&-
    \\
    &&&&\ding{52} &-&-&-&-&34.44
    \\
    \hline
    \ding{52}&\ding{52}&&&&32.56 & 34.39 &- &- &-
    \\
    \ding{52}&&\ding{52}&&&32.45 & - &41.99 &- &-
    \\
    &\ding{52}&\ding{52}&&&- & 34.33 &41.83 &- &-
    \\
    &&&\ding{52}&\ding{52}&- & - &- &33.88 &34.32
    \\
    \hline
     \ding{52}&\ding{52}& \ding{52}&&&31.43 & 34.22 &41.15 &- &-
    \\
    \ding{52}&\ding{52}& &\ding{52}&&31.32 & 34.13 &- &31.45 &-
    \\
    \ding{52}&\ding{52}& &&\ding{52}&31.40 & 34.19 &- &- &34.02
    \\
    \hline
    \ding{52}&\ding{52}& \ding{52}&\ding{52}&&31.23 & 34.19 &40.88 &31.43 &-
    \\
    \ding{52}&\ding{52}& \ding{52}&&\ding{52}&31.25 & 34.07 &40.92 &- &33.96
    \\
    \ding{52}&\ding{52}& &\ding{52}&\ding{52}&31.22 & 34.15 &- &31.48 &33.97
    \\
    \hline
   \ding{52} &\ding{52} &\ding{52} &\ding{52} &\ding{52} &31.11&34.01&40.86&31.41&33.76
    \\
    \hline
\end{tabular}}
\end{table}

\subsubsection{Effect of different combinations of restoration tasks}
To analyze the impact of task composition in the proposed all-in-one restoration framework, we conduct extensive ablation experiments under different combinations of degradation types, including R, S, H, B, N. The corresponding quantitative results are reported in Table~\ref{tb:ablta}.

The model achieves the best performance when trained on a single degradation type, indicating that task-specific optimization provides the most specialized restoration capability for individual corruption distributions. 
When multiple degradation types are jointly introduced, a clear performance trade-off can be observed across individual tasks. In general, as the number of joint tasks increases, the performance on each specific degradation slightly decreases, reflecting the inherent optimization conflict in learning heterogeneous restoration objectives within a unified model.

Despite this trade-off, carefully designed task combinations can still benefit certain restoration tasks through shared degradation priors. For instance, the combination of rain, snow, haze, and noise  achieves stronger deraining and desnowing performance compared to configurations that replace haze or snow with blur. This suggests that degradations such as haze and noise, which primarily affect global intensity and low-frequency statistics, may provide complementary supervisory signals that facilitate more transferable feature learning.
In contrast, motion blur introduces spatially structured distortions that differ significantly from other degradation types. As shown in Table~\ref{tb:ablta}, configurations involving blur tend to slightly reduce performance on other tasks, indicating potential optimization conflicts between spatially variant and globally distributed degradations.

\begin{table} 
\centering 
\caption{Plug-and-play replacement results obtained by substituting the original Mamba blocks in existing restoration architectures with the proposed Multi-$\tau$ Liquid-Mamba.} 
\label{tab:plug} 
\resizebox{\linewidth}{!}{ 
    \begin{tabular}{c|cccc} 
    \hline 
    Method & $\triangle$ Params(M) & $\triangle$ Flops(G)& PSNR & $\triangle$ PSNR 
    \\ \hline \hline 
    ACL~\cite{aclgu2025acl}&-& -& 32.78 & - 
    \\ 
    Replace with Multi-$\tau$ Liquid-Mamba &+0.013 &+2.74 & 33.99 & +1.21 
    \\ \hline 
    MambaIRv2~\cite{guo2025mambairv2}&-&- & 32.56 & - 
    \\ 
    Replace with Multi-$\tau$ Liquid-Mamba &+0.045 &+3.62 &33.92 & +1.36 
    \\ \hline 
    TSPMamba~\cite{TSP-MambaZhou_2025_CVPR} &-&-& 31.79 & - 
    \\ 
    Replace with Multi-$\tau$ Liquid-Mamba &+0.012 & +2.56 &33.65 & +1.86
    \\ \hline 
    \end{tabular}} 
\end{table}

\begin{figure}
    \centerline{\includegraphics[width=1\linewidth]{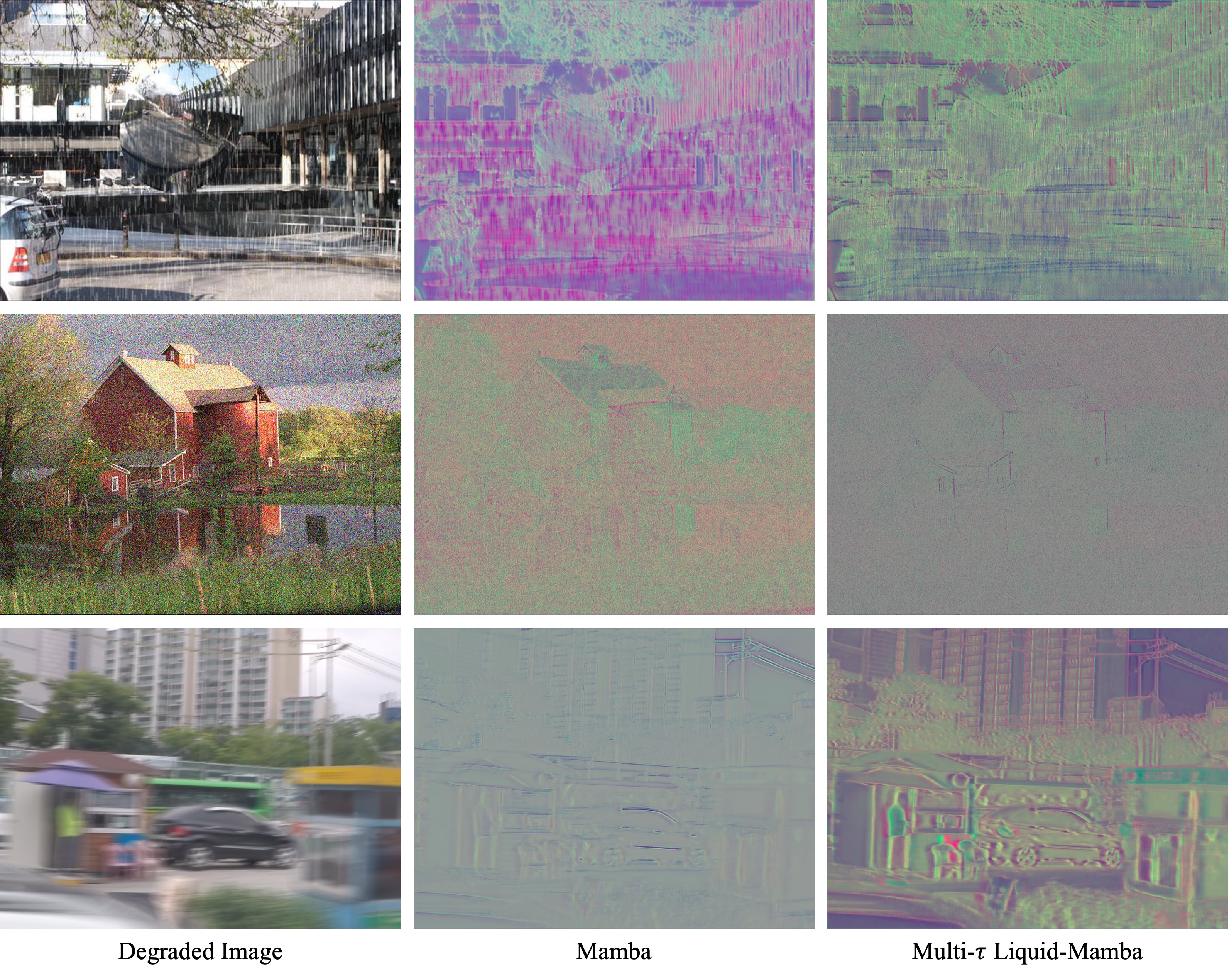}}
	\caption{Visualization of restoration responses under different methods. From left to right: degraded inputs, results of the baseline Mamba, and results of the proposed Multi-$\tau$ Liquid-Mamba. The proposed method adaptively adjusts multi-scale temporal dynamics across different degradation types, leading to more effective restoration and clearer structural recovery.}
 \label{fig:fea} 
\end{figure}

\subsubsection{Plug-and-Play Integration into Existing Architectures}

To evaluate the generality and transferability of the proposed Multi-$\tau$ Liquid-Mamba, we conduct a plug-and-play replacement study on several representative Mamba-based restoration frameworks, including ACL~\cite{aclgu2025acl}, MambaIRv2~\cite{guo2025mambairv2}, and TSPMamba~\cite{TSP-MambaZhou_2025_CVPR}. Specifically, we replace the original Mamba blocks in each model with our Multi-$\tau$ Liquid-Mamba, while keeping all other architectural components and training settings unchanged.

As shown in Table~\ref{tab:plug}, consistent performance improvements are observed across all evaluated architectures. For ACL, the replacement yields a significant improvement from 32.78 dB to 33.99 dB, achieving a gain of +1.21 dB with only marginal increases in computational cost (+0.013M parameters and +2.74G FLOPs). Similarly, MambaIRv2 benefits from a larger gain, improving from 32.56 dB to 33.92 dB (+1.36 dB), accompanied by a slight overhead of +0.045M parameters and +3.62G FLOPs. These results demonstrate that introducing multi-timescale liquid dynamics can effectively enhance the representation capacity of selective state-space models without altering their overall architectural design.
We further validate the effectiveness of our method on TSPMamba, a recent task-specific restoration framework. After replacing its original Mamba modules with Multi-$\tau$ Liquid-Mamba, the performance improves from 31.79 dB to 33.65 dB, corresponding to a substantial gain of +1.86 dB, with only minimal additional computational cost (+0.012M parameters and +2.56G FLOPs).

To further investigate the effectiveness of the proposed Multi-$\tau$ Liquid-Mamba, we visualize the features across different degradation types in Figure~\ref{fig:fea}. 
As shown in the figure, the baseline Mamba exhibits noticeable limitations when handling heterogeneous degradations. In particular, its restoration results tend to suffer from residual artifacts and insufficient structure recovery. 
In contrast, the proposed Multi-$\tau$ Liquid-Mamba consistently produces clearer and more faithful reconstructions across all cases. By introducing multiple temporal evolution rates, the model is able to adaptively adjust its restoration behavior according to the underlying degradation characteristics. As a result, different degradation patterns  are effectively handled within a unified framework, leading to improved structural preservation and reduced artifacts.

\subsubsection{Effect of the multi-$\tau$ design}

\begin{table}
    \centering
      \caption{Effect of the multi-$\tau$ design.}
    \label{tab:multi-de}
    \begin{tabular}{c|cc}
    \hline
        Method & PSNR & $\triangle$ PSNR
         \\
         \hline \hline
          Multi-$\tau$& 34.23 & -
         \\
          Single-$\tau$  & 33.05 & -1.18
         \\
        w/o $\tau$ & 32.59 & -1.64
          \\
         \hline
    \end{tabular}
\end{table}

\begin{figure*}
    \centerline{\includegraphics[width=1\linewidth]{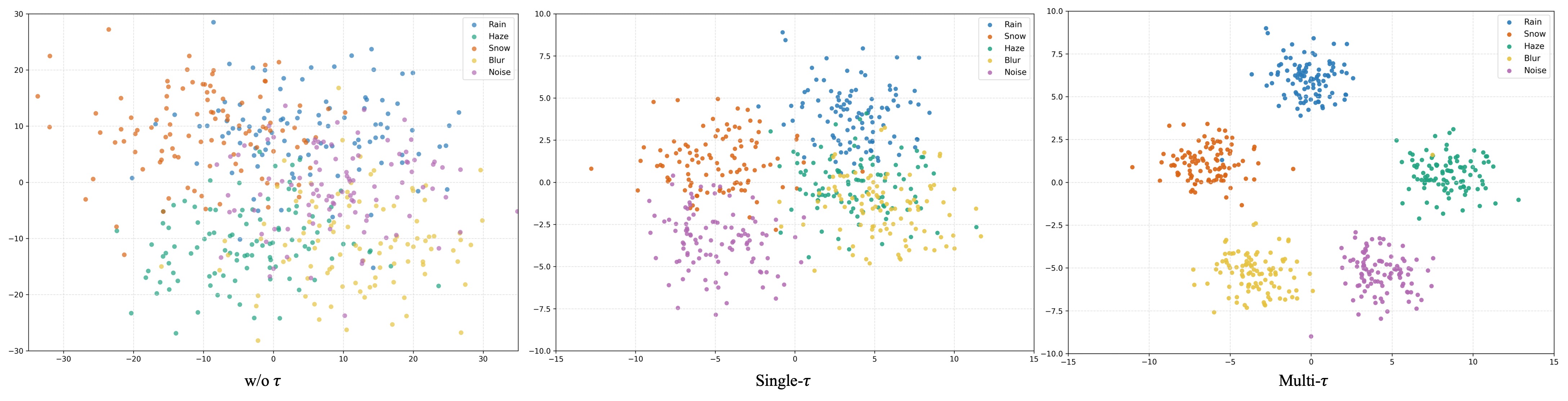}}
	\caption{T-SNE visualization of learned feature distributions under different $\tau$ designs (w/o $\tau$, Single-$\tau$, and Multi-$\tau$).}
 \label{fig:tau} 
\end{figure*}

The effectiveness of the proposed multi-$\tau$ design is evaluated in Table~\ref{tab:multi-de}. Specifically, we compare the complete Multi-$\tau$ Liquid-Mamba with two variants: \textit{i)} Single-$\tau$, which employs only a single adaptive time constant, and \textit{ii)} w/o $\tau$, which removes the liquid timescale modeling and degenerates to the conventional Mamba-style state transition mechanism.
As shown in Table~\ref{tab:multi-de}, the complete Multi-$\tau$ design achieves the best performance. When the multiple timescale mechanism is reduced to a single adaptive time constant, the performance drops to 33.05 dB, resulting in a decrease of 1.18 dB. This indicates that a single temporal scale is insufficient to model the diverse degradation characteristics encountered in all-in-one image restoration. 
Furthermore, removing the $\tau$-based liquid dynamics entirely leads to a more pronounced performance degradation. Compared with the full model, this variant suffers a decrease of 1.64 dB, demonstrating the importance of adaptive timescale modulation in the proposed framework. Without dynamic memory control, the state evolution becomes less flexible and cannot effectively adapt to degradation-dependent restoration requirements.

To further evaluate the effectiveness of the proposed multi-$\tau$ design, we visualize the learned feature distributions using t-SNE, as shown in Figure~\ref{fig:tau}. As can be observed, when the $\tau$-based temporal modeling is removed (w/o $\tau$), features corresponding to different degradation types are heavily entangled. The resulting feature space exhibits substantial overlap and lacks a clear clustering structure, indicating limited capability in distinguishing heterogeneous degradation patterns without explicit temporal dynamics.
Introducing a Single-$\tau$ design leads to a more organized feature distribution, where preliminary clustering behavior emerges among different degradation types. Nevertheless, noticeable overlaps remain between several degradation categories, suggesting that a single evolution timescale is insufficient to capture the diverse restoration dynamics required for multiple degradations.
In contrast, the proposed Multi-$\tau$ design produces well-separated and compact clusters for all degradation types. Different degradations occupy distinct regions in the feature space, exhibiting substantially improved intra-class compactness and inter-class separability. These results demonstrate that modeling multiple temporal evolution scales enables the network to adaptively capture degradation-specific dynamics, resulting in more discriminative and disentangled feature representations.

\begin{table}
    \centering
      \caption{Effect of different multi-$\tau$ state fusion strategies.}
    \label{tab:multi-de2}
    \begin{tabular}{c|cc}
    \hline
        Method & PSNR & $\triangle$ PSNR
         \\
         \hline \hline
         Add & 33.78 & -
         \\
         Concat + Conv  & 33.91 & +0.13
         \\
        Adaptive (Ours) & 34.23 & +0.45
          \\
         \hline
    \end{tabular}
\end{table}

The proposed Multi-$\tau$ Liquid-Mamba maintains multiple state trajectories operating at different temporal scales. Therefore, an effective fusion mechanism is required to aggregate the information from different dynamic branches. Table~\ref{tab:multi-de2} compares three fusion strategies, including direct summation (\emph{Add}), channel concatenation followed by convolution (\emph{Concat + Conv}), and the proposed adaptive fusion mechanism.
As shown in Table~\ref{tab:multi-de2}, directly summing the multi-$\tau$ states yields a PSNR of 33.78 dB. Although this strategy introduces negligible computational overhead, it treats all temporal branches equally and ignores their degradation-dependent importance. Replacing it with channel concatenation followed by a convolutional projection improves the performance to 33.91 dB, indicating that learnable interactions among different temporal scales are beneficial for restoration.
The proposed adaptive fusion strategy achieves the best performance, reaching 34.23 dB PSNR. Compared with direct summation and \emph{Concat + Conv}, it provides improvements of 0.45 dB and 0.32 dB, respectively. This gain can be attributed to the degradation-aware weighting mechanism, which dynamically adjusts the contribution of each temporal branch according to the input content.

\begin{table}[h]
\centering
\caption{Ablation study on the number of timescales $M$.}
\label{tab:tau_ablation}
\begin{tabular}{c|ccc}
\hline
$M$  & PSNR   & $\triangle$ PSNR & $\triangle$ FLOPs(G)
\\
\hline\hline
1 & 33.05  & - & -\\ 
2  & 33.68 & +0.63 & +1.25 \\
3  & 33.94 &+0.89 & +2.11\\
4  & 34.23 &+1.18&+2.74\\
6  & 34.17 &+1.12&+3.82 \\
\hline
\end{tabular}
\end{table}

To investigate the impact of the proposed multi-timescale mechanism, we conduct an ablation study on the number of timescales $M$ in Multi-$\tau$ Liquid-Mamba. We evaluate the model under different settings $M \in \{1, 2, 3, 4, 6\}$, while keeping all other training and architectural configurations unchanged.
As shown in Table~\ref{tab:tau_ablation}, the model with a single timescale ($M=1$) degenerates to a standard liquid-state Mamba variant and yields the lowest performance, suggesting that a single temporal evolution pattern is insufficient for modeling complex and spatially varying degradations. As the number of timescales increases, the restoration performance improves consistently, demonstrating that the proposed multi-timescale design effectively enhances the model's ability to capture both rapidly varying local details and slowly evolving global contextual information.
The performance reaches its optimum at $M=4$, achieving 34.23 dB PSNR. This result indicates that four timescales provide a favorable balance between representation capacity and redundancy across different temporal dynamics. When $M$ is further increased to 6, the performance exhibits a slight saturation and marginal decline, which may be attributed to the increased correlation among timescale branches and the diminishing benefit of additional dynamical components. Therefore, $M=4$ is adopted as the default setting throughout this paper.
From an efficiency perspective, the computational overhead introduced by additional timescales is relatively small, as evidenced by the marginal increase in FLOPs with increasing $M$. This observation suggests that the performance improvements primarily stem from enhanced dynamical modeling capability rather than a substantial increase in model capacity, highlighting the effectiveness of the proposed multi-timescale liquid dynamics.

\begin{figure}
    \centerline{\includegraphics[width=1\linewidth]{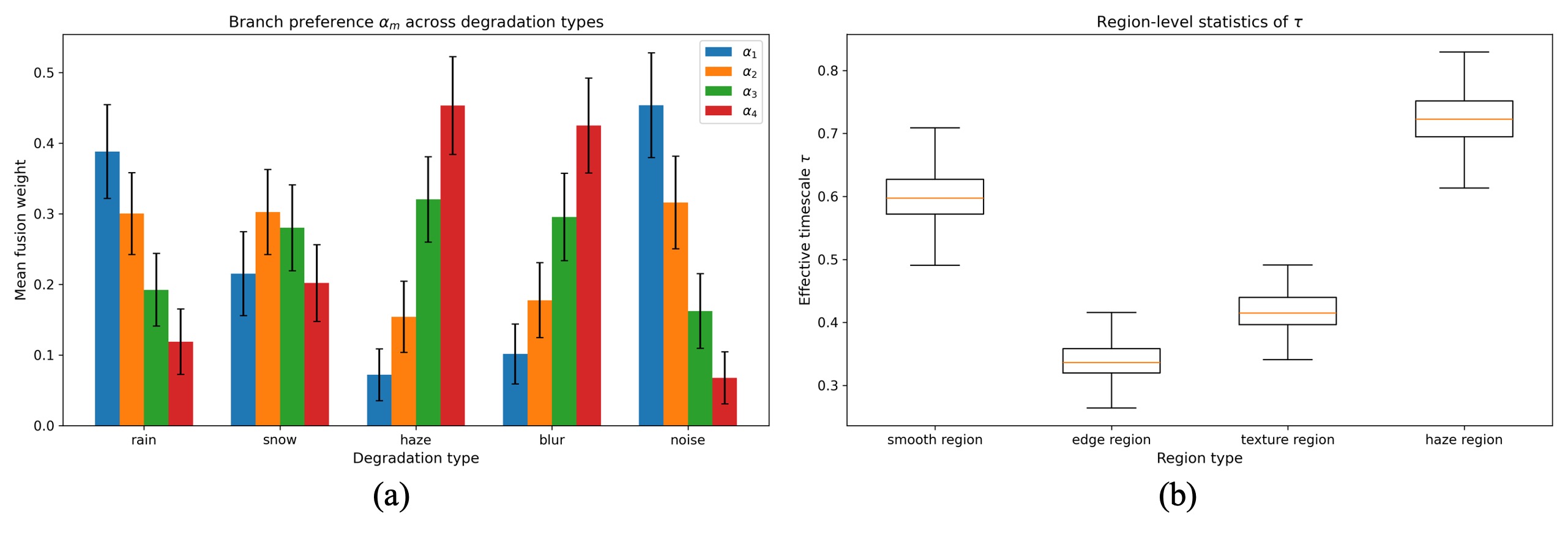}}
	\caption{Visualization and statistical analysis of degradation-dependent multi-timescale dynamics.
(a) Average fusion weights ($\alpha_m$) of different dynamical branches under representative degradation types.
(b) Region-level statistics of the branch-level scalar timescale for visualization over smooth, edge, texture, and haze regions.}
 \label{fig:tau_alp} 
\end{figure}

To further investigate whether the proposed Multi-\(\tau\) Liquid-Mamba indeed learns degradation-dependent state-evolution patterns, we analyze the learned adaptive fusion weights \(\alpha_m\) and liquid timescales
As shown in Figure~\ref{fig:tau_alp}(a), different degradation types exhibit distinct branch-selection patterns. Rain and noise tend to assign larger weights to short-timescale branches, suggesting that these locally dominated degradations require faster state evolution to suppress high-frequency corruptions and recover local details. In contrast, haze and blur show higher responses on long-timescale branches, indicating that the model relies more on slowly evolving dynamics and broader contextual aggregation to handle spatially extended degradation structures. Snow exhibits a more balanced distribution over multiple branches, which is consistent with its mixed degradation characteristics involving both local occlusion and global visibility distortion.
We further analyze the spatial behavior of the learned dynamics by computing \(\tau\) over different image regions, including smooth, edge, texture, and haze regions. As shown in Figure~\ref{fig:tau_alp}(b), smooth and haze regions exhibit larger effective timescales, implying that the model adopts longer memory horizons to preserve global consistency and model slowly varying degradation patterns. By contrast, edge and texture regions have smaller timescales, suggesting that the model favors faster state transitions for high-frequency structures and fine-grained details. This region-level difference indicates that the proposed module is not only sensitive to degradation categories, but also adaptively adjusts its state-evolution behavior according to local image content.

These observations provide direct evidence that MLMIR does not rely on a fixed transition mechanism. Instead, the proposed Multi-\(\tau\) Liquid-Mamba learns to dynamically select different temporal branches and memory horizons according to both degradation type and spatial content. This adaptive multi-timescale behavior explains why MLMIR can effectively handle heterogeneous degradations in the all-in-one restoration setting.

\section{Discussion}

\subsection{Design Principles}

The proposed Multi-$\tau$ Liquid-Mamba is built upon three core design principles. First, we introduce adaptive state transition dynamics, where continuous-time liquid neural modeling is incorporated into selective state space models to improve adaptability under spatially varying degradations. Second, we adopt a multi-timescale formulation, enabling the model to simultaneously capture fast-changing local structures and slowly varying global contextual dependencies through multiple learnable time constants. Third, we ensure plug-and-play compatibility, such that the proposed module can be seamlessly integrated into existing Mamba-based restoration architectures without modifying their overall pipeline, making it a general and reusable component for diverse image restoration tasks.

\subsection{Computational Complexity}

We analyze the computational complexity of Multi-$\tau$ Liquid-Mamba with respect to the input sequence length $L$, feature dimension $D$, state dimension $N$, and the number of timescales $M$. In standard selective state space models, the dominant computation arises from the selective scan operation, which can be implemented in $\mathcal{O}(L \cdot D \cdot N)$ complexity due to linear-time recurrent state updates.

In our method, each time step maintains $M$ parallel liquid state trajectories with different time constants $\{\tau_m\}_{m=1}^M$. The state update for each branch follows the same linear recurrent form as the original Mamba-style scan, resulting in a per-branch complexity of $\mathcal{O}(L \cdot D \cdot N)$. Aggregating all branches yields $\mathcal{O}(M \cdot L \cdot D \cdot N)$.
In addition, the computation of the timescale gating parameters and liquid modulation terms (e.g., $\tau_m$, $\alpha_m$, and input-dependent transition modulation) introduces an extra projection cost of $\mathcal{O}(L \cdot D^2)$, which is consistent with standard linear projection layers in Mamba-style blocks. Therefore, the total complexity of the proposed Multi-$\tau$ Liquid-Mamba is $\mathcal{O}(M \cdot L \cdot D \cdot N + L \cdot D^2)$.
Since $M$ is a small constant (e.g., $M \in [2,6]$ in practice) and $N \ll D$ in typical configurations, the proposed method preserves the linear scaling property $\mathcal{O}(L)$ with respect to sequence length $L$.
Therefore, Multi-$\tau$ Liquid-Mamba retains the efficiency advantage of selective state space models while introducing only marginal overhead for multi-timescale dynamics.

\subsection{Limitations and Future Work}

Although the proposed method consistently improves performance across different restoration architectures, several limitations remain. First, the introduction of multiple timescale hyperparameters increases the model design space. Second, the method does not explicitly incorporate physical degradation priors, which may limit interpretability under extremely complex or unseen degradation conditions. Third, the current formulation is primarily validated on single-image restoration tasks, and its behavior in temporally consistent scenarios such as video restoration remains unexplored.

Future research can be extended in several directions. A promising avenue is to integrate explicit physical degradation modeling with liquid state space dynamics to further enhance robustness and interpretability. Another direction is to extend the proposed framework to video and multi-frame restoration tasks, where continuous-time dynamics may naturally benefit temporal modeling. Additionally, developing adaptive mechanisms to automatically determine the optimal number of timescales could further improve scalability and reduce manual design effort.

\section{Conclusion}

In this work, we propose Multi-$\tau$ Liquid-Mamba, a novel adaptive state space modeling framework that integrates continuous-time liquid neural dynamics into selective state space models to address the limited adaptability of existing Mamba-based image restoration methods under spatially varying degradations. By introducing input-dependent multi-timescale state transitions, the proposed method enables flexible and content-aware temporal evolution, significantly enhancing the expressiveness of state transition modeling while preserving the linear complexity and efficiency of Mamba. Importantly, Multi-$\tau$ Liquid-Mamba is designed as a plug-and-play module that can be seamlessly integrated into a wide range of existing restoration architectures, consistently improving performance across both task-aligned and all-in-one restoration settings. Built upon this design, we further develop MLMIR, a hierarchical encoder--decoder network that jointly captures global contextual dependencies and local structural details. Extensive experiments demonstrate that our method achieves state-of-the-art performance on multiple restoration benchmarks and consistently improves diverse Mamba-based backbones, validating its generality.

\bibliographystyle{IEEEtran}
\bibliography{ref}

\vfill

\end{document}